%% file: main.tex
\documentclass{article}
\usepackage{iclr2026_conference}

\usepackage[T1]{fontenc}
\usepackage{mathptmx}

\input{math_commands.tex}

\usepackage{caption}
\usepackage{subcaption}
\usepackage{graphicx}
\usepackage{wrapfig}
\graphicspath{{figures/}}
\usepackage{booktabs}
\usepackage{enumitem}
\usepackage{multirow}

\usepackage{amssymb}
\usepackage{amsmath}

\usepackage{algorithmic}
\usepackage{array}
\usepackage{fancyvrb}
\usepackage{pgfplots}
\usepackage{hyperref}
\usepackage{url}
\usepackage[frozencache]{minted}
\usepackage{float}
\usepackage{adjustbox}
\usepackage{tikz}
\usepackage{tabularx}
\usetikzlibrary{decorations}
\usepackage{rotating}
\usepackage{verbatim}
\usepackage{xcolor}
\definecolor{darkblue}{RGB}{0,0,139}
\hypersetup{colorlinks=true, citecolor=darkblue}

\title{Automatically Generating Hard Math Problems from Hypothesis-Driven Error Analysis}

\author{
Jiayu Fu$^{1}$, Mourad Heddaya$^{1}$, Chenhao Tan$^{1}$\\[0.3em]
$^{1}$Department of Computer Science, University of Chicago, Chicago, IL 60637, USA\\[0.3em]
\texttt{\{fujiayu, mourad, chenhao\}@uchicago.edu}
}

\iclrfinalcopy %
\begin{document}

\maketitle

\begin{abstract}
Numerous math benchmarks exist to evaluate LLMs' mathematical capabilities. However, most involve extensive manual effort and are difficult to scale. Consequently, they cannot keep pace with LLM development or easily provide new instances to mitigate overfitting. Some researchers have proposed automatic benchmark generation methods, but few focus on identifying the specific math concepts and skills on which LLMs are error-prone, and most can only generate category-specific benchmarks. To address these limitations, we propose a new math benchmark generation pipeline that uses AI-generated hypotheses to identify the specific math concepts and skills that LLMs struggle with, and then generates new benchmark problems targeting these weaknesses. Experiments show that hypothesis accuracy positively correlates with the difficulty of the generated problems: problems generated from the most accurate hypotheses reduce Llama-3.3-70B-Instruct's accuracy to as low as 45\%, compared to 77\% on the original MATH benchmark. Furthermore, our pipeline is highly adaptable and can be applied beyond math to explore a wide range of LLM capabilities, making it a valuable tool for investigating how LLMs perform across different domains.
\end{abstract}

\input{sections/01-introduction}
\input{sections/02-related-works}
\input{sections/03-generation-pipeline}
\input{sections/04-evaluating-problem-generation}
\input{sections/05-general-discussion}
\input{sections/06-conclusion}

\bibliographystyle{iclr2026_conference}

\input{main.bbl}
\newpage
\appendix
\input{sections/07-appendix}

\end{document}

%% file: math_commands.tex
\usepackage{amsmath,amsfonts,bm}

\def\eqref#1{equation~\ref{#1}}

\def\1{\bm{1}}

\DeclareMathAlphabet{\mathsfit}{\encodingdefault}{\sfdefault}{m}{sl}
\SetMathAlphabet{\mathsfit}{bold}{\encodingdefault}{\sfdefault}{bx}{n}

%% file: sections/01-introduction.tex
\section{Introduction}
\label{sec:introduction}
Evaluating the mathematical reasoning of LLMs relies on benchmarks, yet most existing benchmarks are manually constructed and curated. This manual process does not scale, creating two persistent problems. First, because benchmark problems are static and publicly available, LLMs frequently encounter them during training---sometimes inadvertently, sometimes by design---leading to overfitted evaluations that overestimate true capability. Efforts to generate fresh instances that resemble the original benchmark~\citep{zhang2024accessing} still rely on manual construction and therefore inherit the same bottleneck. Second, LLMs are evolving rapidly, but manual benchmark creation cannot keep pace: existing benchmarks quickly become obsolete, while new benchmarks targeting emerging capabilities are slow to appear.

To overcome these limitations, recent work has explored LLM-based benchmark generation. While faster than manual construction, current automatic methods share several shortcomings: (i)~they are often restricted to a single problem format such as multiple-choice or simple QA; (ii)~they do not identify the specific mathematical concepts and skills on which LLMs are weakest, and therefore cannot target generated problems toward those weaknesses; and (iii)~they tend to be domain-specific, covering only a narrow slice of mathematics rather than adapting across topics or beyond math entirely.

We introduce a benchmark generation pipeline that addresses all three gaps. Our pipeline uses Hypogenic~\citep{zhou2024hypothesisgenerationlargelanguage}, an LLM-based hypothesis generator, to analyze problems that a target LLM consistently fails and produce hypotheses about the mathematical concepts and skills underlying those failures. These hypotheses then guide the generation of new problems designed to probe the identified weaknesses. The pipeline is lightweight and adaptable: by modifying the hypothesis prompt, it can investigate non-mathematical factors that affect LLM performance (e.g., problem wording or solution length) or generate benchmarks in domains outside of mathematics.

We evaluate the pipeline on the MATH benchmark~\citep{hendrycksmath2021} using Llama-3.3-70B-Instruct~\citep{meta-llama_llama-3.3-70b-instruct_2024} as the target model, testing across five levels of mathematical concept granularity and three hypothesis-generating LLMs. Our experiments show a positive correlation between hypothesis accuracy and the difficulty of the resulting problems: problems generated from the most accurate hypotheses reduce the target model's accuracy to as low as 45\%, compared to 77\% on the original MATH benchmark. We also find that granularity matters---redundant or overlapping concept categories degrade both hypothesis quality and downstream problem quality.

Our contributions are as follows:
\begin{enumerate}
    \item A generation pipeline that identifies mathematical concepts and skills on which LLMs are weak and produces targeted benchmark problems that are substantially more challenging than general-purpose benchmarks.
    \item An empirical analysis showing that the granularity of concept categorization significantly affects both hypothesis accuracy and generated problem quality, with redundant categories degrading performance.
\end{enumerate}

%% file: sections/02-related-works.tex
\section{Related Works}
\label{sec:related-works}
Most existing math benchmarks, such as GSM8k~\citep{cobbe2021gsm8k} and MATH~\citep{hendrycksmath2021}, require extensive manual effort to construct. Even recent benchmarks rely on substantial human involvement: \citet{glazer2024frontiermath} enlist expert mathematicians to create and verify a dataset of challenging problems, and \citet{chernyshev2024u} manually curate a university-level benchmark. Because human-in-the-loop construction is slow and difficult to scale, it struggles to keep pace with LLM progress or to refresh problems quickly enough to prevent overfitting. This has motivated a growing body of work on partially or fully automatic benchmark generation.

\paragraph{Extending existing benchmarks.}
Several methods generate new problems by transforming or mimicking existing ones. \citet{huang2024datagen} train an LLM-based pipeline to produce new instances that match the length, semantic embedding, and diversity of a source benchmark. \citet{o2025mathemagic} modify problems with counterfactual rules to test induction, deduction, and overfitting tendencies. Other work focuses on increasing difficulty: \citet{wang2024benchmark} evolve the complexity of existing benchmarks through multi-agent interaction, and \citet{zhang2024darg} extract and perturb reasoning graphs to produce higher-complexity samples.

\paragraph{Prompt-based generation.}
Rather than extending existing data, some approaches generate benchmarks directly from prompts. \citet{yuan2025llm} develop BenchMaker, which automatically elaborates a user-provided prompt into detailed multiple-choice questions. \citet{shashidhar2025yourbench} introduce YourBench, a framework that generates QA or MCQ problems using an input document as reference. A limitation of these prompt-based approaches is that they require either seed data or reference documents, making them difficult to apply in domains that lack such resources.

\paragraph{Tool-integrated generation.}
A third line of work combines multiple LLM capabilities in the generation process. \citet{shah2025aiassistedgenerationdifficultmath} extract and categorize mathematical skills from the MATH benchmark, then prompt an LLM to generate new problems by combining these skills. \citet{peng2025proof2hybrid} build an automatic proof-benchmark generator for algebraic geometry that produces problems resilient to guessing and superficial pattern matching.

\paragraph{Hypothesis generation with LLMs.}
Our pipeline builds on Hypogenic~\citep{zhou2024hypothesisgenerationlargelanguage}, a framework for generating and evaluating natural-language hypotheses from labeled data using LLMs. Hypogenic iteratively proposes hypotheses, scores them against held-out examples, and returns the most accurate ones. We repurpose this machinery to hypothesize which mathematical concepts and skills underlie an LLM's failures, then use those hypotheses to guide targeted problem generation.

Among existing automatic generation methods, few attempt to identify the specific concepts and skills on which LLMs are most error-prone, and none investigate how the granularity of concept categorization affects generation quality. Our work addresses both gaps.

%% file: sections/03-generation-pipeline.tex
\section{Generation Pipeline}
\label{sec:generation-pipeline}
\begin{figure}[H]
    \centering
    \includegraphics[width=0.9\linewidth]{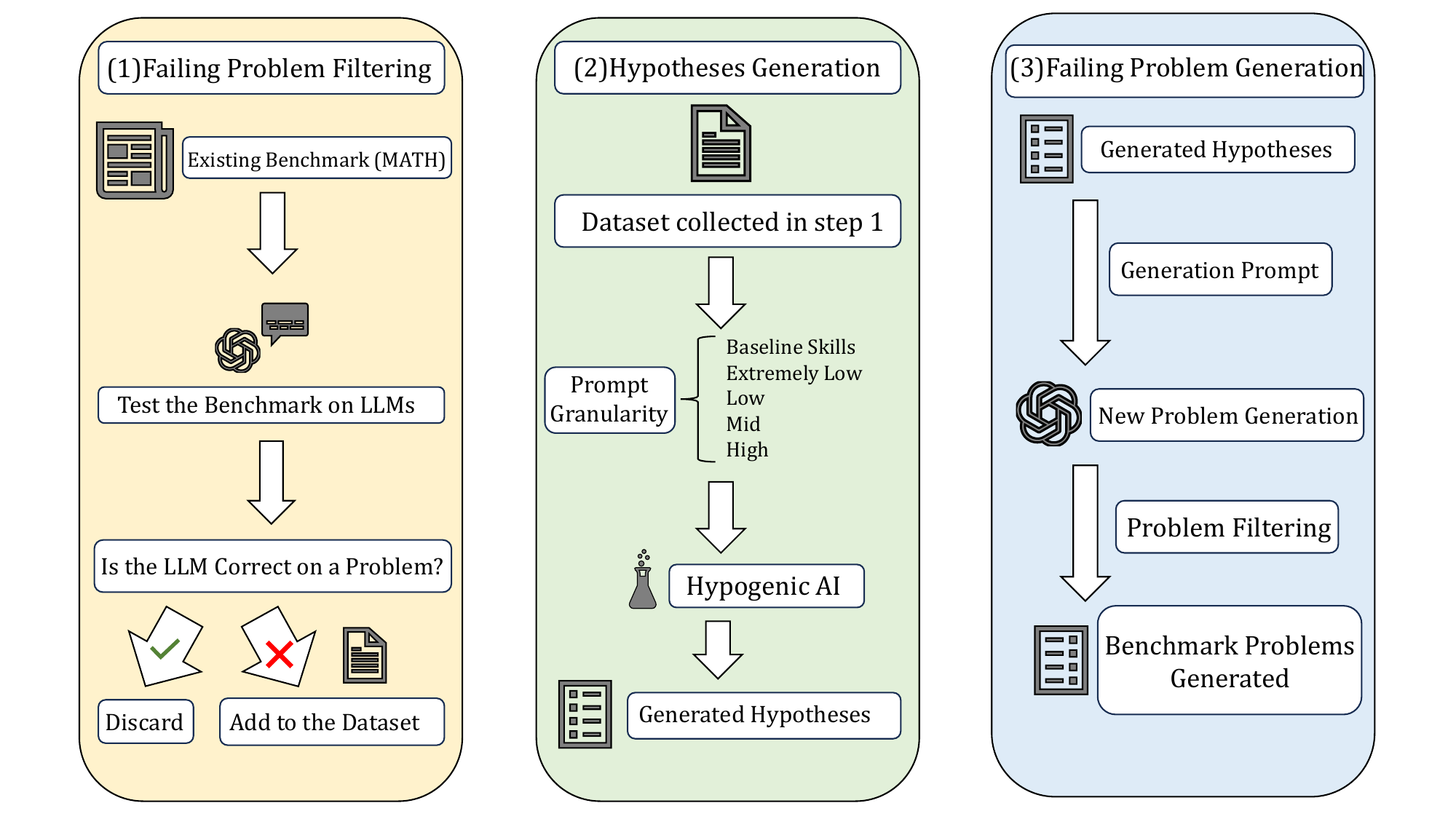}
    \caption{\small{Overview of the three-stage generation pipeline: (1)~filter problems that the target LLM consistently fails, (2)~generate hypotheses about the concepts and skills underlying those failures, and (3)~generate new problems guided by the hypotheses.}}
    \label{fig:generation-pipeline}
\end{figure}
Our pipeline generates challenging math problems through three stages (Figure~\ref{fig:generation-pipeline}):
\begin{enumerate}
    \item \textbf{Failing Problem Filtering:}
    Test the target LLM on an existing benchmark and extract problems it consistently answers incorrectly.
    \item \textbf{Hypothesis Generation:}
    Use an LLM to generate hypotheses for which mathematical concepts and skills underlie the failures.
    \item \textbf{Challenging Problem Generation:}
    Generate new problems that target the identified weaknesses, guided by the hypotheses from Stage~2.
\end{enumerate}
\subsection{Failing Problem Filtering}
\label{subsection:Failing Problem Filtering}
We evaluate Llama-3.3-70B-Instruct~\citep{meta-llama_llama-3.3-70b-instruct_2024} on the MATH benchmark using the model's recommended decoding parameters (temperature\,=\,0.6, top\_p\,=\,0.9, top\_k\,=\,40, repetition\_penalty\,=\,1.2). Each problem is attempted five times; we retain only those that the model answers incorrectly on every attempt. Because these failures persist across all five trials, they are unlikely to result from sampling variance and more likely reflect systematic weaknesses in the model's mathematical reasoning. The resulting set of consistently failed problems forms the input to the hypothesis generation stage.
\subsection{Hypothesis Generation}
\label{subsection: Hypothesis Generation}
We run Hypogenic~\citep{zhou2024hypothesisgenerationlargelanguage} on the failed-problem dataset from Stage~1. Hypogenic proposes natural-language hypotheses about which mathematical concepts and skills are associated with the LLM's failures, then scores each hypothesis by its \emph{accuracy}: the fraction of samples in the dataset whose labels (correct/incorrect) are consistent with the hypothesis. We retain the top fifteen hypotheses per configuration for downstream evaluation (listed in Appendix~\ref{appendix:sample_hypotheses}).

\begin{figure}[htbp]
    \centering
    \includegraphics[width=0.85\linewidth]{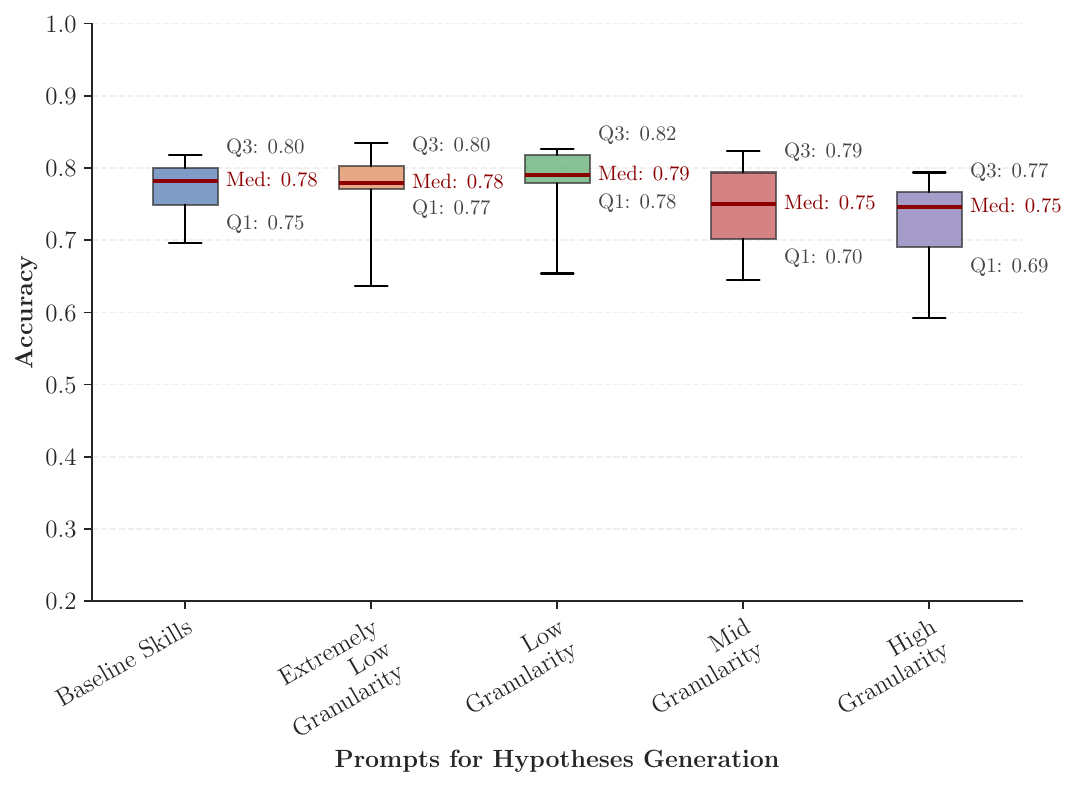}
    \caption{\small{Hypothesis accuracy distributions across granularity levels (using GPT-4.1-mini during hypotheses generation). The low-granularity prompt achieves the highest median and quartile accuracies. Accuracy increases from extremely low to low granularity, then decreases as granularity increases further.}}
    \label{fig:gpt4.1mini-boxplot}
\end{figure}

To investigate how the granularity of concept categorization affects hypothesis quality, we design five prompt variants (Appendix~\ref{appendix:prompts}), each providing Hypogenic with a different taxonomy of mathematical concepts and skills. Four taxonomies---\emph{extremely low}, \emph{low}, \emph{mid}, and \emph{high} granularity---were constructed by applying LLM-based extraction to the MATH benchmark at increasing levels of specificity. A fifth \emph{baseline} taxonomy uses the full skill list extracted by \citet{shah2025aiassistedgenerationdifficultmath}. We also compare three LLMs as the backbone for Hypogenic: GPT-4o-mini, GPT-4.1-mini, and Qwen3-14B (with thinking disabled). Among these, GPT-4.1-mini produces the most accurate hypotheses, and the low-granularity prompt yields the best results overall. Figure~\ref{fig:gpt4.1mini-boxplot} shows the hypothesis accuracy distributions across granularities for GPT-4.1-mini; results for the other two models appear in Appendix~\ref{appendix:granularity_performance}.

Because the generated hypotheses capture patterns in the concepts and skills that the target LLM struggles with, they serve as natural guides for generating challenging problems. We use them as part of the generation prompt in Stage~3.
\subsection{Challenging Problem Generation}
\label{subsection:Challenging Problem Generation}
We construct generation prompts by combining the hypotheses from Stage~2 with the mathematical concepts and skills present in the MATH benchmark (see Appendix~\ref{appendix:problem_generation_prompts} for full prompts). Since GPT-4.1-mini produces the most accurate hypotheses, we use its outputs for all problem generation. The generating LLM is Llama-3.3-70B-Instruct---the same model used in Stage~1---to control for model-specific variation throughout the pipeline (Hypogenic is the sole exception).

Because the generated problems target areas where LLMs are weak, the generating model itself is prone to producing flawed outputs. We therefore apply a multi-stage filtering process: we remove problems that are invalid, incorrect, incomplete, or ambiguous, and additionally exclude proof-style questions to standardize the output format and simplify evaluation. Answer keys are derived using DeepSeek-R1, GPT-o3, and GPT-5, then verified through cross-model agreement and manual validation. Sample generated problems for each granularity are shown in Appendix~\ref{appendix:sample_generated_math_problems}.

%% file: sections/04-evaluating-problem-generation.tex
\section{Evaluating Problem Generation}
\label{sec:evaluating-problem-generation}
\label{section:04evaluating-problem-generation}
We evaluate the quality of the generated problems by measuring how often the target LLM fails on them. For each granularity level, we select the two highest-accuracy hypotheses (from those listed in Appendix~\ref{appendix:sample_hypotheses}) and generate problems following the procedure in Section~\ref{subsection:Challenging Problem Generation}.
\subsection{Experimental Setup}
Problem generation uses Llama-3.3-70B-Instruct with slightly elevated diversity settings (temperature\,=\,1.0, top\_p\,=\,0.9, top\_k\,=\,50, repetition\_penalty\,=\,1.05) to encourage variety. For each hypothesis, we randomly sample 20 problems from the filtered pool, with answer keys derived as described in Section~\ref{subsection:Challenging Problem Generation}.

To maintain consistency, we evaluate the generated problems using the same model and decoding configuration as in Stage~1 (Llama-3.3-70B-Instruct; temperature\,=\,0.6, top\_p\,=\,0.9, top\_k\,=\,40, repetition\_penalty\,=\,1.2). This avoids confounds from model-specific differences in problem-solving ability, since we observe that larger or more capable models tend to produce substantially harder problems under our prompting.

\begin{table}[h!]
\centering
\renewcommand{\arraystretch}{1.4}
\begin{tabularx}{\linewidth}{>{\centering\arraybackslash}p{2.8cm} X >{\centering\arraybackslash}p{2.5cm}}
\toprule
\textbf{Granularity} & \textbf{Hypothesis} & \textbf{Llama-3.3 Solve Rate} \\
\midrule

& The LLM is likely to fail in problems requiring calculation and conversion skills. & 90\% \\
\cmidrule(l){2-3}
\multirow{-2}{2.8cm}{\centering Baseline Skills}
& The LLM shows difficulty on problems involving coordinate geometry and transformation skills together with graph understanding and interpretation. & 95\% \\
\midrule

& The LLM is likely to fail on problems involving the combination of Geometry and Algebra. & 70\% \\
\cmidrule(l){2-3}
\multirow{-2}{2.8cm}{\centering Extremely Low Granularity}
& The LLM is likely to fail on problems involving both Prealgebra and Algebra. & 90\% \\
\midrule

& The LLM is likely to fail on problems involving modular arithmetic, divisibility, and integer properties. & 45\% \\
\cmidrule(l){2-3}
\multirow{-2}{2.8cm}{\centering Low Granularity}
& The LLM is likely to fail on problems involving spatial reasoning and geometric theorem application. & 60\% \\
\midrule

& The LLM is more error-prone on problems involving function evaluation/composition. & 90\% \\
\cmidrule(l){2-3}
\multirow{-2}{2.8cm}{\centering Mid Granularity}
& The LLM is more likely to fail on problems requiring the use of linear \& systems concepts. & 95\% \\
\midrule

& The LLM is likely to fail in problems involving function evaluation and basic transformations. & 55\% \\
\cmidrule(l){2-3}
\multirow{-2}{2.8cm}{\centering High Granularity}
& The LLM is likely to fail on problems involving integer arithmetic ($+,-,\times,\div$). & 95\% \\
\bottomrule

\end{tabularx}
\caption{Selected hypotheses and Llama-3.3-70B-Instruct solve rates on problems generated from each hypothesis, across granularity levels.}
\label{tab:hypotheses_and_accuracies}
\end{table}

\subsection{Experimental Results}
Table~\ref{tab:hypotheses_and_accuracies} reports the two selected hypotheses per granularity and Llama-3.3-70B-Instruct's solve rate on each 20-problem set; Figure~\ref{fig:generated_problem_performance} visualizes these results.

\textbf{Hypothesis accuracy predicts problem difficulty.} Across all five granularity levels, the model consistently performs worse on problems generated from the highest-accuracy hypothesis than on those from the second-highest. This confirms that Hypogenic's accuracy scores meaningfully reflect the degree to which a hypothesis captures the model's weaknesses: higher-accuracy hypotheses yield harder problems.

\begin{figure}[htbp]
    \centering
    \includegraphics[width=0.8\linewidth]{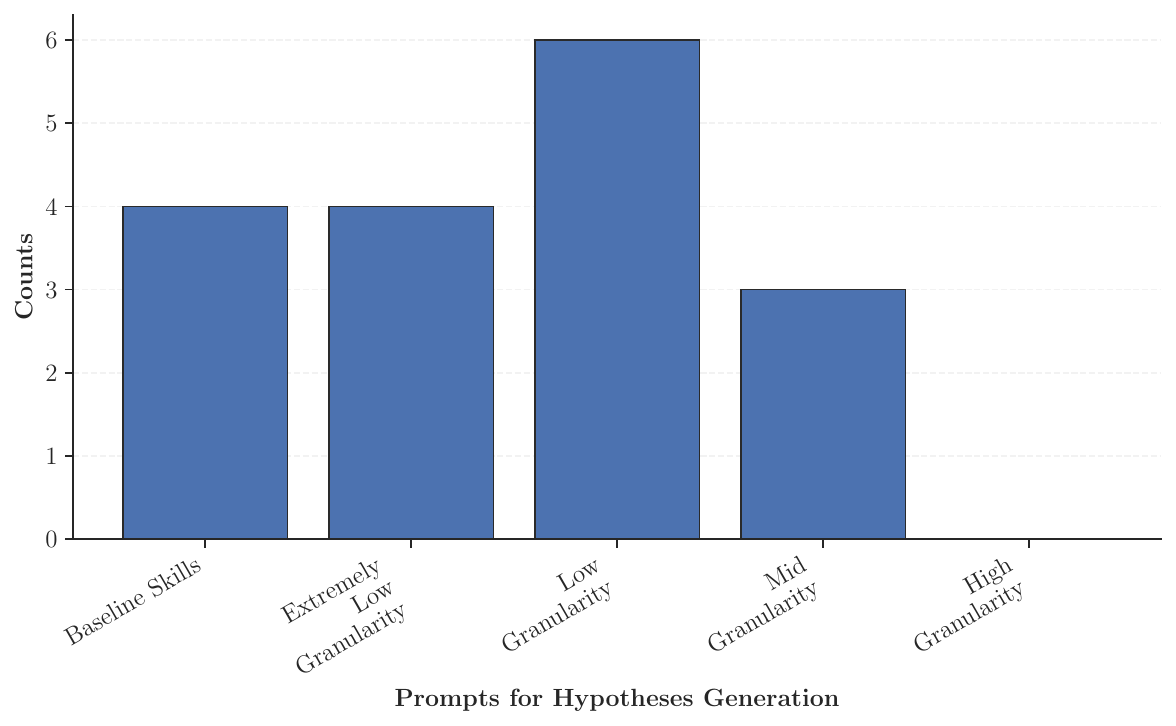}
    \caption{Number of Hypotheses with Accuracies Over 0.8 Using GPT-4.1-mini under Different Prompts for Hypotheses Generation}
    \label{fig:high accuracy hypotheses count}
\end{figure}

\begin{figure}[htbp]
    \centering
    \includegraphics[width=0.8\linewidth]{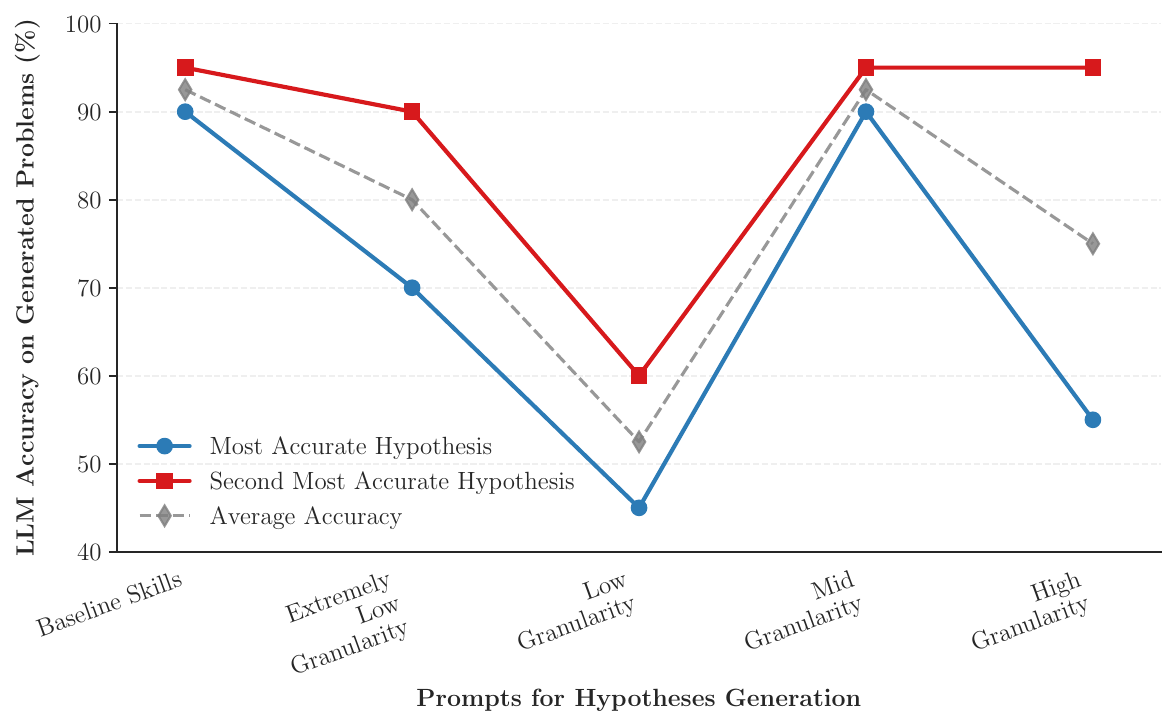}
    \caption{\small{Llama-3.3-70B-Instruct solve rates on generated problems under hypotheses generated with different prompts. The model's solve rate on the original MATH benchmark is 77\%~\citep{meta-llama_llama-3.3-70b-instruct_2024}. Solve rate decreases from extremely low to low granularity, then increases from low to high, mirroring the trend in the number of high-accuracy hypotheses (Figure \ref{fig:high accuracy hypotheses count}).}}
    \label{fig:generated_problem_performance}
\end{figure}

\textbf{Granularity affects generation quality.} The trend in solve rates across granularities mirrors the trend in the number of high-accuracy hypotheses (Figure~\ref{fig:high accuracy hypotheses count}). The low-granularity prompt produces the most hypotheses with accuracy above 0.8, and problems generated from low-granularity hypotheses are also the most challenging---reducing the model's solve rate to as low as 45\%. Under baseline and mid granularities, even the best hypothesis produces problems on which the model scores at or above its 77\% MATH benchmark accuracy, suggesting that overly coarse categories lack specificity while overly fine ones constrain generation creativity.

\textbf{Redundant categories hurt performance.} Problems generated from baseline-skill hypotheses are the least challenging overall (Figure~\ref{fig:generated_problem_performance}), indicating that redundant and overlapping concept categories degrade both hypothesis quality and the difficulty of the resulting problems.

%% file: sections/05-general-discussion.tex
\section{General Discussion}
\label{sec:General Discussion}

\subsection{Advantages over Existing Methods}
Our experiments confirm that hypothesis-guided generation produces problems that are meaningfully harder for the target LLM, and that hypothesis accuracy is a reliable predictor of problem difficulty. Compared to existing automatic generation approaches, our pipeline offers several advantages. First, it explicitly identifies the concepts and skills on which an LLM is weakest, rather than generating problems uniformly or relying on surface-level difficulty heuristics. Second, it accounts for the granularity of concept categorization, a factor that, to our knowledge, no prior work has investigated, and shows that redundant or overlapping categories degrade generation quality. Third, the pipeline requires no manually curated seed set beyond an existing benchmark and involves minimal human intervention.

Beyond targeted benchmark construction, the generated problems can shed light on how LLMs interpret mathematical concepts differently from humans, potentially explaining why models sometimes solve advanced problems while failing on elementary ones. Because the pipeline's behavior is controlled entirely through the Hypogenic prompt, it can be adapted to explore non-mathematical factors (such as problem wording, solution length, or the number of concepts per problem) or extended to domains outside mathematics.

\subsection{Limitations}

\paragraph{Small evaluation sample.} Due to computational constraints, we evaluate only 20 problems per hypothesis. At this sample size, each individual error corresponds to a 5 percentage-point shift in solve rate, making the results sensitive to noise. Scaling to larger problem sets would yield more precise estimates and clearer trends across granularity levels.

\paragraph{Generator capability.} Llama-3.3-70B-Instruct serves as both the target and the generator. We observe that the model produces more flawed problems when guided by high-accuracy hypotheses, precisely the areas where it is weakest. Although we filter these problems, generator limitations may still reduce the quality ceiling of the benchmark, particularly for the most targeted hypotheses.

\paragraph{Correlation vs.\ causation.} The hypotheses identify statistical associations between concept labels and failures, but these associations are not necessarily causal. An LLM may fail on a problem labeled ``modular arithmetic'' for reasons unrelated to modular arithmetic itself:
\begin{enumerate}
    \item \emph{Sensitivity to wording:} the model may struggle with specific phrasings, and paraphrasing the same problem could yield a correct answer.
    \item \emph{Long-context degradation:} problems requiring extended reasoning chains may cause the model to lose coherence, independent of the mathematical content.
    \item \emph{Confounding skills:} problems nominally testing one concept often involve auxiliary skills (e.g., solving a geometry problem via systems of equations), and the true source of error may lie in the auxiliary skill rather than the labeled one.
\end{enumerate}
Addressing this limitation is a natural extension of the pipeline: by modifying the Hypogenic prompt to generate hypotheses about non-content factors (e.g., wording complexity, solution length), the same framework can disentangle concept-level weaknesses from other sources of failure.

%% file: sections/06-conclusion.tex
\section{Conclusion}
\label{sec:conclusion}
We presented an automatic math benchmark generation pipeline that uses LLM-powered hypothesis generation to identify mathematical concepts and skills on which a target LLM is weakest, then generates problems that specifically target those weaknesses. Our experiments show that hypothesis accuracy correlates with the difficulty of the resulting problems, with the low-granularity prompt producing the most accurate hypotheses and the hardest generated benchmarks. The pipeline requires only an existing benchmark as input and minimal human oversight, and can be adapted to investigate non-mathematical failure factors or extended to other domains by modifying the hypothesis prompt. In future work, we plan to scale evaluation to larger problem sets, test additional target models, and explore prompts that disentangle concept-level weaknesses from confounding factors such as problem wording and solution length.

%% file: sections/07-appendix.tex
\section{Hypogenic Prompts}
\label{appendix:prompts}
\subsection{Baseline Skill Prompt}
\begin{minted}[fontsize=\normalsize, breaklines, breakanywhere]{yaml}
prompt_templates:
  observations: |
    The math problem: ${problems}
    The LLM's answer to the problem: ${answers}
    The correctness of the answer: ${label}.

  batched_generation:
    system: |-
      You are a professional math teacher and educational researcher.
      Given a set of math problems and LLM-generated answers, we want to generate hypotheses that predicts LLMs are more error-prone on which particular math skills/concepts (or combinations of math skills/concepts)
      In other words, we want to understand what kinds of problem are associated with correct or wrong labels, what kind of problem makes LLM more likely to fail,
      what kind of math concept or combination of math concepts make LLM likely to fail the problem. Here is a form of math concepts involved in the problems:
        "absolute_value_skills",
        "algebra_and_equations",
        "algebraic_expression_skills",
        "algebraic_manipulation_and_equations",
        "algebraic_manipulation_skills",
        "algebraic_skills",
        "area_calculation_skills",
        "arithmetic_operations",
        "arithmetic_sequences",
        "arithmetic_skills",
        "average_calculations",
        "base_conversion",
        "basic_arithmetic",
        "basic_arithmetic_operations",
        "basic_trigonometry",
        "calculating_and_understanding_combinations",
        "calculation_and_conversion_skills",
        "calculus",
        "calculus_skills",
        "circle_geometry_skills",
        "circles",
        "combinatorial_mathematics",
        "combinatorial_operations_and_basic_arithmetic",
        "combinatorics_and_probability_skills",
        "combinatorics_knowledge",
        "complex_number_manipulation_and_operations",
        "complex_number_operations",
        "complex_number_skills",
        "complex_numbers",
        "complex_numbers_related_skills",
        "coordinate_geometry_and_transformation_skills",
        "coordinate_systems",
        "counting_and_number_theory",
        "counting_principals",
        "distance_and_midpoint_skills",
        "division_and_remainders",
        "exponent_and_root_skills",
        "exponentials_and_logarithms",
        "exponentiation",
        "exponentiation_rules",
        "factorials_and_prime_factorization",
        "factoring_skills",
        "factorization",
        "fractions_and_decimals",
        "function_composition_and_transformation",
        "function_composition_skills",
        "function_skills",
        "geometric_relations",
        "geometric_sequence_skills",
        "geometric_series_comprehension",
        "geometry",
        "geometry_and_space_calculation",
        "geometry_triangle_properties",
        "graph_and_geometry_skills",
        "graph_understanding_and_interpretation",
        "greatest_common_divisor_calculations",
        "inequality_skills",
        "inequality_solving_and_understanding",
        "logarithmic_and_exponential_skills",
        "matrix_operations",
        "modular_arithmetic",
        "multiplication_and_division",
        "number_manipulation",
        "number_theory",
        "number_theory_and_arithmetic_operations",
        "number_theory_skills",
        "other_geometric_skills",
        "parametric_equations",
        "perimeter_and_area",
        "permutation_and_combinations",
        "polynomial_skills",
        "prime_number_recognition_and_properties",
        "prime_number_theory",
        "probability_and_combinatorics",
        "probability_concepts_and_calculations",
        "properties_and_application_of_exponents",
        "pythagorean_skills",
        "quadratic_equation_skills",
        "quadratic_equations_and_solutions",
        "ratio_and_proportion",
        "ratio_and_proportion_skills",
        "recurrence",
        "recursive_functions_and_sequences",
        "sequence_analysis",
        "sequence_and_series_analysis_skills",
        "sequence_and_series_skills",
        "sequences",
        "sequences_series_and_summation",
        "simplification_and_basic_operations",
        "solving_equations",
        "solving_inequalities",
        "solving_linear_equation",
        "solving_system_of_equations",
        "summation_and_analysis_of_series",
        "three_dimensional_geometry",
        "triangle_geometry_skills",
        "trigonometry_skills",
        "understanding_and_application_of_functions",
        "understanding_and_applying_combinatorics_concepts",
        "understanding_and_applying_floor_and_ceiling_functions",
        "understanding_and_manipulation_of_rational_functions",
        "understanding_and_utilizing_infininte_series",
        "understanding_circle_properties_and_algebraic_manipulation",
        "understanding_ellipse_properties",
        "understanding_logarithmic_properties_and_solving_equations",
        "understanding_of_fractions",
        "vector_operations"

      Using !!!ONLY!!! the given math concepts in the form, !!!DO NOT!!! include any guessings or conditions other than the concepts in the form!!!
      !!!Do NOT!!! propose conditions or special particularities. For example, a hypothesis like "The LLM is likely to fail on a concept, especially/particularly when the problem is...(some special conditions)" or "The LLM is likely to fail on a concept because of (some condition)" is not allowed!
      For example: "The LLM is likely to fail on problems that require function evaluation and transformations, especially when dealing with composite functions and inverses." is a BAD hypothesis that doesn't follow the previous requirement.
      Another example: "The LLM is likely to fail on (some math concept) when the problem is complex" is also a BAD hypothesis that doesn't follow the instructions. 'complex' is very vague and is a condition that is NOT in the math concept. It will NOT be accepted.
      !!!Do NOT!!! propose hypotheses only from a particular step in the problem! For example, a hypothesis like "The LLM is likely to fail on a concept at a (some step in problem solution)" is not allowed!
      !!!Do NOT!!! propose any reasons behind the failures! For example, a hypothesis like "The LLM is likely to fail on a concept, because/due to ... (some reason)" is not allowed!
      Again, Use !!!Only!!! the given math concepts in the form! Only means the hypotheses can only contain the concepts in the form! and no other things allowed!
      Using anything other than the math concepts in the form will NOT be accepted and should NOT be proposed!
      These hypotheses should identify specific patterns that occur across the provided problems and LLM-generated answers.
      please propose ${num_hypotheses} possible hypotheses.
      These hypotheses should identify specific patterns that occur across the provided problems and LLM-generated answers.
      You should check carefully the specific solving steps by the LLM, and consider which particular
      step and which particular math concept/skill did the LLM make mistake on.
      When proposing hypotheses, generate half hypotheses using a single math concept, and generate the other half by combining two or more math concepts.
      Again, use ONLY and VERBATIMLY the provided math concepts from the list

      Each hypothesis should be:
        A hypothesis about what particular math makes the LLM to fail.

      Generate them in the format of 1. [hypothesis], 2. [hypothesis], ... ${num_hypotheses}. [hypothesis].
      The hypotheses should analyze what particular math concept(s) are associated with correctness or error.

    user: |-
      We have seen some math problems and LLM-generated answers:
      ${observations}
      Please generate hypotheses that are useful for predicting which particular math concept and solution step does the LLM likely to make mistakes on.
      Propose ${num_hypotheses} possible hypotheses. Generate them in the format of 1. [hypothesis], 2. [hypothesis], ... ${num_hypotheses}. [hypothesis].
      Proposed hypotheses:

  inference:
    system: |-
      You are a professional math teacher and your job is to determine whether a given answer to a math problem is correct or wrong.
      From past experience, you have learned that LLMs are more likely to fail on certain math concepts (or combination of math concepts).
      You need to determine whether the learned pattern applies to the current problem and answer, and then make your prediction.
      Give your final answer in the format of "Final answer: answer", where the answer is either "correct" or "wrong".

    user: |-
      Our learned pattern: ${hypothesis}
      A math problem and its answer are the following:
      Problem: "${problems}"
      Answer: "${answers}"

      Given the pattern you learned above, decide whether the answer is correct or wrong.
      Think step by step.
      First step: Consider if the pattern can be applied to the answer.
      Second step: Based on the pattern, is this answer correct or wrong?
      Final step: give your final answer in the format of "Final answer: answer"

  multiple_hypotheses_inference:
    system: |-
      You are a professional math teacher and your job is to determine whether a given answer to a math problem is correct or wrong.
      From past experience, you have learned that LLMs are more likely to fail on certain math concepts (or combination of math concepts).
      You need to determine whether these patterns apply to the current problem and answer, and then make your prediction.
      Give your final answer in the format of "Final answer: answer", where the answer is either "correct" or "wrong".

    user: |-
      Our learned patterns: ${hypotheses}
      A math problem and its answer are the following:
      Problem: "${problems}"
      Answer: "${answers}"

      Given the patterns you learned above, decide whether the answer is correct or wrong.
      Think step by step.
      First step: Think about which pattern(s) can be applied to the answer.
      Second step: Based on the patterns, is this answer correct or wrong?
      Final step: give your final answer in the format of "Final answer: answer"
\end{minted}
\subsection{Extremely Low Granularity Prompt}
\begin{minted}[fontsize=\normalsize, breaklines, breakanywhere]{yaml}
prompt_templates:
  observations: |
    The math problem: ${problems}
    The LLM's answer to the problem: ${answers}
    The correctness of the answer: ${label}.

  batched_generation:
    system: |-
      You are a professional math teacher and educational researcher.
      Given a set of math problems and LLM-generated answers, we want to generate hypotheses that predicts LLMs are more error-prone on which particular math skills/concepts (or combinations of math skills/concepts)
      In other words, we want to understand what kinds of problem are associated with correct or wrong labels, what kind of problem makes LLM more likely to fail,
      what kind of math concept or combination of math concepts make LLM likely to fail the problem. Here is a form of math concepts involved in the problems:
      1. Prealgebra
      2. Algebra
      3. Number Theory
      4. Counting & Probability
      5. Geometry
      6. Precalculus
      7. Advanced Auxiliary Topics

      Using !!!ONLY!!! the given math concepts in the form, !!!DO NOT!!! include any guessings or conditions other than the concepts in the form!!!
      !!!Do NOT!!! propose conditions or special particularities. For example, a hypothesis like "The LLM is likely to fail on a concept, especially/particularly when the problem is...(some special conditions)" or "The LLM is likely to fail on a concept because of (some condition)" is not allowed!
      For example: "The LLM is likely to fail on problems that require function evaluation and transformations, especially when dealing with composite functions and inverses." is a BAD hypothesis that doesn't follow the previous requirement.
      Another example: "The LLM is likely to fail on (some math concept) when the problem is complex" is also a BAD hypothesis that doesn't follow the instructions. 'complex' is very vague and is a condition that is NOT in the math concept. It will NOT be accepted.
      !!!Do NOT!!! propose hypotheses only from a particular step in the problem! For example, a hypothesis like "The LLM is likely to fail on a concept at a (some step in problem solution)" is not allowed!
      !!!Do NOT!!! propose any reasons behind the failures! For example, a hypothesis like "The LLM is likely to fail on a concept, because/due to ... (some reason)" is not allowed!
      Again, Use !!!Only!!! the given math concepts in the form! Only means the hypotheses can only contain the concepts in the form! and no other things allowed!
      Using anything other than the math concepts in the form will NOT be accepted and should NOT be proposed!
      These hypotheses should identify specific patterns that occur across the provided problems and LLM-generated answers.
      please propose ${num_hypotheses} possible hypotheses.
      These hypotheses should identify specific patterns that occur across the provided problems and LLM-generated answers.
      You should check carefully the specific solving steps by the LLM, and consider which particular
      step and which particular math concept/skill did the LLM make mistake on.
      When proposing hypotheses, generate half hypotheses using a single math concept, and generate the other half by combining two or more math concepts.
      Again, use ONLY and VERBATIMLY the provided math concepts from the list

      Each hypothesis should be:
        A hypothesis about what particular math makes the LLM to fail.

      Generate them in the format of 1. [hypothesis], 2. [hypothesis], ... ${num_hypotheses}. [hypothesis].
      The hypotheses should analyze what particular math concept(s) are associated with correctness or error.

    user: |-
      We have seen some math problems and LLM-generated answers:
      ${observations}
      Please generate hypotheses that are useful for predicting which particular math concept and solution step does the LLM likely to make mistakes on.
      Propose ${num_hypotheses} possible hypotheses. Generate them in the format of 1. [hypothesis], 2. [hypothesis], ... ${num_hypotheses}. [hypothesis].
      Proposed hypotheses:

  inference:
    system: |-
      You are a professional math teacher and your job is to determine whether a given answer to a math problem is correct or wrong.
      From past experience, you have learned that LLMs are more likely to fail on certain math concepts (or combination of math concepts).
      You need to determine whether the learned pattern applies to the current problem and answer, and then make your prediction.
      Give your final answer in the format of "Final answer: answer", where the answer is either "correct" or "wrong".

    user: |-
      Our learned pattern: ${hypothesis}
      A math problem and its answer are the following:
      Problem: "${problems}"
      Answer: "${answers}"

      Given the pattern you learned above, decide whether the answer is correct or wrong.
      Think step by step.
      First step: Consider if the pattern can be applied to the answer.
      Second step: Based on the pattern, is this answer correct or wrong?
      Final step: give your final answer in the format of "Final answer: answer"

  multiple_hypotheses_inference:
    system: |-
      You are a professional math teacher and your job is to determine whether a given answer to a math problem is correct or wrong.
      From past experience, you have learned that LLMs are more likely to fail on certain math concepts (or combination of math concepts).
      You need to determine whether these patterns apply to the current problem and answer, and then make your prediction.
      Give your final answer in the format of "Final answer: answer", where the answer is either "correct" or "wrong".

    user: |-
      Our learned patterns: ${hypotheses}
      A math problem and its answer are the following:
      Problem: "${problems}"
      Answer: "${answers}"

      Given the patterns you learned above, decide whether the answer is correct or wrong.
      Think step by step.
      First step: Think about which pattern(s) can be applied to the answer.
      Second step: Based on the patterns, is this answer correct or wrong?
      Final step: give your final answer in the format of "Final answer: answer"
\end{minted}
\subsection{Low Granularity Prompt}
\begin{minted}[fontsize=\normalsize, breaklines, breakanywhere]{yaml}
prompt_templates:
  observations: |
    The math problem: ${problems}
    The LLM's answer to the problem: ${answers}
    The correctness of the answer: ${label}.

  batched_generation:
    system: |-
      You are a professional math teacher and educational researcher.
      Given a set of math problems and LLM-generated answers, we want to generate hypotheses that predicts LLMs are more error-prone on which particular math skills/concepts (or combinations of math skills/concepts)
      In other words, we want to understand what kinds of problem are associated with correct or wrong labels, what kind of problem makes LLM more likely to fail,
      what kind of math concept or combination of math concepts make LLM likely to fail the problem. Here is a form of math concepts involved in the problems:
      | Math Concepts/Skills                                 |
      | ---------------------------------------------------- |
      | Expression manipulation, equation solving            |
      | Modular arithmetic, divisibility, integer properties |
      | Combinatorics, probability modeling                  |
      | Spatial reasoning, theorem application               |
      | Sequences, function analysis                         |
      | Multi-step reasoning, deduction, diagram use         |
      | parabolas, ellipses, hyperbolas, graph of GCD        |

      Using !!!ONLY!!! the given math concepts in the form, !!!DO NOT!!! include any guessings or conditions other than the concepts in the form!!!
      !!!Do NOT!!! propose conditions or special particularities. For example, a hypothesis like "The LLM is likely to fail on a concept, especially/particularly when the problem is...(some special conditions)" or "The LLM is likely to fail on a concept because of (some condition)" is not allowed!
      For example: "The LLM is likely to fail on problems that require function evaluation and transformations, especially when dealing with composite functions and inverses." is a BAD hypothesis that doesn't follow the previous requirement.
      Another example: "The LLM is likely to fail on (some math concept) when the problem is complex" is also a BAD hypothesis that doesn't follow the instructions. 'complex' is very vague and is a condition that is NOT in the math concept. It will NOT be accepted.
      !!!Do NOT!!! propose hypotheses only from a particular step in the problem! For example, a hypothesis like "The LLM is likely to fail on a concept at a (some step in problem solution)" is not allowed!
      !!!Do NOT!!! propose any reasons behind the failures! For example, a hypothesis like "The LLM is likely to fail on a concept, because/due to ... (some reason)" is not allowed!
      Again, Use !!!Only!!! the given math concepts in the form! Only means the hypotheses can only contain the concepts in the form! and no other things allowed!
      Using anything other than the math concepts in the form will NOT be accepted and should NOT be proposed!
      These hypotheses should identify specific patterns that occur across the provided problems and LLM-generated answers.
      please propose ${num_hypotheses} possible hypotheses.
      These hypotheses should identify specific patterns that occur across the provided problems and LLM-generated answers.
      You should check carefully the specific solving steps by the LLM, and consider which particular
      step and which particular math concept/skill did the LLM make mistake on.
      When proposing hypotheses, generate half hypotheses using a single math concept, and generate the other half by combining two or more math concepts.
      Again, use ONLY and VERBATIMLY the provided math concepts from the list

      Each hypothesis should be:
        A hypothesis about what particular math makes the LLM to fail.

      Generate them in the format of 1. [hypothesis], 2. [hypothesis], ... ${num_hypotheses}. [hypothesis].
      The hypotheses should analyze what particular math concept(s) are associated with correctness or error.

    user: |-
      We have seen some math problems and LLM-generated answers:
      ${observations}
      Please generate hypotheses that are useful for predicting which particular math concept and solution step does the LLM likely to make mistakes on.
      Propose ${num_hypotheses} possible hypotheses. Generate them in the format of 1. [hypothesis], 2. [hypothesis], ... ${num_hypotheses}. [hypothesis].
      Proposed hypotheses:

  inference:
    system: |-
      You are a professional math teacher and your job is to determine whether a given answer to a math problem is correct or wrong.
      From past experience, you have learned that LLMs are more likely to fail on certain math concepts (or combination of math concepts).
      You need to determine whether the learned pattern applies to the current problem and answer, and then make your prediction.
      Give your final answer in the format of "Final answer: answer", where the answer is either "correct" or "wrong".

    user: |-
      Our learned pattern: ${hypothesis}
      A math problem and its answer are the following:
      Problem: "${problems}"
      Answer: "${answers}"

      Given the pattern you learned above, decide whether the answer is correct or wrong.
      Think step by step.
      First step: Consider if the pattern can be applied to the answer.
      Second step: Based on the pattern, is this answer correct or wrong?
      Final step: give your final answer in the format of "Final answer: answer"

  multiple_hypotheses_inference:
    system: |-
      You are a professional math teacher and your job is to determine whether a given answer to a math problem is correct or wrong.
      From past experience, you have learned that LLMs are more likely to fail on certain math concepts (or combination of math concepts).
      You need to determine whether these patterns apply to the current problem and answer, and then make your prediction.
      Give your final answer in the format of "Final answer: answer", where the answer is either "correct" or "wrong".

    user: |-
      Our learned patterns: ${hypotheses}
      A math problem and its answer are the following:
      Problem: "${problems}"
      Answer: "${answers}"

      Given the patterns you learned above, decide whether the answer is correct or wrong.
      Think step by step.
      First step: Think about which pattern(s) can be applied to the answer.
      Second step: Based on the patterns, is this answer correct or wrong?
      Final step: give your final answer in the format of "Final answer: answer"
\end{minted}
\subsection{Mid Granularity Prompt}
\begin{minted}[fontsize=\normalsize, breaklines, breakanywhere]{yaml}
prompt_templates:
  observations: |
    The math problem: ${problems}
    The LLM's answer to the problem: ${answers}
    The correctness of the answer: ${label}.

  batched_generation:
    system: |-
      You are a professional math teacher and educational researcher.
      Given a set of math problems and LLM-generated answers, we want to generate hypotheses that predicts LLMs are more error-prone on which particular math skills/concepts (or combinations of math skills/concepts)
      In other words, we want to understand what kinds of problem are associated with correct or wrong labels, what kind of problem makes LLM more likely to fail,
      what kind of math concept or combination of math concepts make LLM likely to fail the problem. Here is a form of math concepts involved in the problems:

      Math Concepts/Skills
      -Addition/subtraction/multiplication/division (fractions, decimals), PEMDAS, even/odd, factors, multiples, GCD, LCM, absolute value, integer exponents, roots
      -Linear & systems, inequalities, polynomial ops, factoring quadratics, rational expressions, exponents & radicals, absolute value eqs/ineqs, function eval/composition, matrix inverse
      -Prime factorization, divisibility, GCD/LCM, modular arithmetic, Euler's phi function, Chinese remainder, parity
      -Permutations, combinations, binomial expansions, inclusion-exclusion, basic probability types, enumeration
      -Angles, triangle theorems (Pythagorean, similarity, congruence), inradius, polygon angles, circle theorems, area/volume (2D/3D), coordinate formulas, polyhedron metrics
      -Polynomial division, factor theorem, rational functions/asymptotes, nonlinear systems, inequalities
      -Sequences & series, exponential/logarithmic equations, basic trig identities/solutions, function inversion/transformation
      -Conic sections, polynomial GCDs, De-Moivre, calculus (integration, arclength, gradients, divergence, curl, Jacobian, Laplacian), linear algebra (eigenvalues, RREF)

      Using !!!ONLY!!! the given math concepts in the form, !!!DO NOT!!! include any guessings or conditions other than the concepts in the form!!!
      !!!Do NOT!!! propose conditions or special particularities. For example, a hypothesis like "The LLM is likely to fail on a concept, especially/particularly when the problem is...(some special conditions)" or "The LLM is likely to fail on a concept because of (some condition)" is not allowed!
      For example: "The LLM is likely to fail on problems that require function evaluation and transformations, especially when dealing with composite functions and inverses." is a BAD hypothesis that doesn't follow the previous requirement.
      Another example: "The LLM is likely to fail on (some math concept) when the problem is complex" is also a BAD hypothesis that doesn't follow the instructions. 'complex' is very vague and is a condition that is NOT in the math concept. It will NOT be accepted.
      !!!Do NOT!!! propose hypotheses only from a particular step in the problem! For example, a hypothesis like "The LLM is likely to fail on a concept at a (some step in problem solution)" is not allowed!
      !!!Do NOT!!! propose any reasons behind the failures! For example, a hypothesis like "The LLM is likely to fail on a concept, because/due to ... (some reason)" is not allowed!
      Again, Use !!!Only!!! the given math concepts in the form! Only means the hypotheses can only contain the concepts in the form! and no other things allowed!
      Using anything other than the math concepts in the form will NOT be accepted and should NOT be proposed!
      These hypotheses should identify specific patterns that occur across the provided problems and LLM-generated answers.
      please propose ${num_hypotheses} possible hypotheses.
      These hypotheses should identify specific patterns that occur across the provided problems and LLM-generated answers.
      You should check carefully the specific solving steps by the LLM, and consider which particular
      step and which particular math concept/skill did the LLM make mistake on.
      When proposing hypotheses, generate half hypotheses using a single math concept, and generate the other half by combining two or more math concepts.
      Again, use ONLY and VERBATIMLY the provided math concepts from the list

      Each hypothesis should be:
        A hypothesis about what particular math makes the LLM to fail.

      Generate them in the format of 1. [hypothesis], 2. [hypothesis], ... ${num_hypotheses}. [hypothesis].
      The hypotheses should analyze what particular math concept(s) are associated with correctness or error.

    user: |-
      We have seen some math problems and LLM-generated answers:
      ${observations}
      Please generate hypotheses that are useful for predicting which particular math concept and solution step does the LLM likely to make mistakes on.
      Propose ${num_hypotheses} possible hypotheses. Generate them in the format of 1. [hypothesis], 2. [hypothesis], ... ${num_hypotheses}. [hypothesis].
      Proposed hypotheses:

  inference:
    system: |-
      You are a professional math teacher and your job is to determine whether a given answer to a math problem is correct or wrong.
      From past experience, you have learned that LLMs are more likely to fail on certain math concepts (or combination of math concepts).
      You need to determine whether the learned pattern applies to the current problem and answer, and then make your prediction.
      Give your final answer in the format of "Final answer: answer", where the answer is either "correct" or "wrong".

    user: |-
      Our learned pattern: ${hypothesis}
      A math problem and its answer are the following:
      Problem: "${problems}"
      Answer: "${answers}"

      Given the pattern you learned above, decide whether the answer is correct or wrong.
      Think step by step.
      First step: Consider if the pattern can be applied to the answer.
      Second step: Based on the pattern, is this answer correct or wrong?
      Final step: give your final answer in the format of "Final answer: answer"

  multiple_hypotheses_inference:
    system: |-
      You are a professional math teacher and your job is to determine whether a given answer to a math problem is correct or wrong.
      From past experience, you have learned that LLMs are more likely to fail on certain math concepts (or combination of math concepts).
      You need to determine whether these patterns apply to the current problem and answer, and then make your prediction.
      Give your final answer in the format of "Final answer: answer", where the answer is either "correct" or "wrong".

    user: |-
      Our learned patterns: ${hypotheses}
      A math problem and its answer are the following:
      Problem: "${problems}"
      Answer: "${answers}"

      Given the patterns you learned above, decide whether the answer is correct or wrong.
      Think step by step.
      First step: Think about which pattern(s) can be applied to the answer.
      Second step: Based on the patterns, is this answer correct or wrong?
      Final step: give your final answer in the format of "Final answer: answer"
\end{minted}
\subsection{High Granularity Prompt}
\begin{minted}[fontsize=\normalsize, breaklines, breakanywhere]{yaml}
prompt_templates:
  observations: |
    The math problem: ${problems}
    The LLM's answer to the problem: ${answers}
    The correctness of the answer: ${label}.

  batched_generation:
    system: |-
      You are a professional math teacher and educational researcher.
      Given a set of math problems and LLM-generated answers, we want to generate hypotheses that predicts LLMs are more error-prone on which particular math skills/concepts (or combinations of math skills/concepts)
      In other words, we want to understand what kinds of problem are associated with correct or wrong labels, what kind of problem makes LLM more likely to fail,
      what kind of math concept or combination of math concepts make LLM likely to fail the problem. Here is a form of math concepts involved in the problems:
      Math Concepts/Skills
      - Integer arithmetic (+,-,*,/) - Fraction operations (simplify, add/subtract, multiply/divide) - Decimal operations (convert, round, compare) - Order of operations (PEMDAS) - Even/odd determination - Divisibility rules (2,3,4,5,6,8,9,11) - Factors and multiples - GCD and LCM via prime factorization - Absolute value - Integer exponents (e.g., 2^n, 3^n) - Square roots & cube roots (perfect and approximations)
      - One-step, multi-step linear equations - Systems of linear equations (substitution/elimination) - Linear inequalities & graphing solution sets - Simplifying polynomials (combine like terms, distributive) - Factoring polynomials: GCF, quadratics, difference of squares - Quadratic solving: factoring, completing the square, quadratic formula, discriminant interpretation - Rational expressions: simplify, multiply/divide, domain restrictions - Radical expressions: simplify, rationalize denominators - Absolute value equations & inequalities - Function evaluation and basic transformations - Function composition - Graphing linear functions (slope-intercept, point-slope) - Slope, intercept, parallel & perpendicular lines - Matrix basics (2*2 inverses, determinants)
      - Prime identification & prime factorization - Fundamental theorem of arithmetic - Divisibility tests (2-11) - GCD/LCM via Euclidean algorithm - Modular arithmetic: congruences, addition/multiplication mod n - Solving linear congruences - Euler's phi function - Chinese remainder theorem - Parity arguments (even/odd reasoning)
      - Factorials (n!) - Permutations: P(n,k) - Combinations: C(n,k), Pascal's triangle patterns - Distinguish permutations vs combinations - Binomial identities & expansion (e.g., (a+b)^n coefficients) - Inclusion-exclusion principle - Basic probability: P(A), conditional (P(A|B)), complement rule, independent vs dependent events - Probability trees and compound events - Counting by cases / systematic enumeration
      - Angle relations: vertical, alternate interior, supplementary, complementary - Triangle properties: Pythagorean theorem, area/perimeter, classification (isosceles, equilateral, right) - Triangle similarity & congruence (AA, SAS, SSS) - Inradius/exradius formulas via area - Polygon properties: interior/exterior angle sums - Circle properties: arc length, sector area, chords, inscribed angles, tangents, central angles - Geometric constructions & relationships (e.g., inscribed shapes) - Area: triangles, quadrilaterals, circles, sectors - Volume & surface area: prisms, cylinders, cones, spheres - Coordinate geometry: distance formula, midpoint, slope, line equations, intercepts - Analytical geometry: shifts, intersections, slopes in coordinate plane - Polyhedron basics (e.g., face/vertex counts via Euler's formula)
      - Polynomial long division & synthetic division - Factor theorem & Remainder theorem - Finding polynomial zeroes/roots - Rational functions: domain, graph behavior, asymptotes - Systems mixing linear and quadratic equations - Quadratic & higher-degree inequalities (sign analysis, test intervals) - Complex algebraic manipulations of expressions/inequalities
      - Sequences & series: arithmetic (a_n=a_1+(n-1)d) & geometric (a_n=a_1*r^n); sum formulas - Binomial theorem and summation identities - Exponential functions and laws (b^n) - Logarithmic functions, properties/log rules, log equations - Exponential/log equations solving (change of base, convert) - Trigonometry: unit circle fundamentals, sin/cos/tan, right-triangle & standard-angle values - Basic trig identities (Pythagorean, double-angle, co-function) - Solving trig equations (within domain restrictions) - Inverse trigonometric functions (arcsin, arccos, arctan) - Function properties: domains, ranges, transformations (shifts/stretches), inverses, composite functions
      - Conic sections: properties of parabolas, ellipses, hyperbolas - Polynomial GCD computations - Complex numbers & De Moivre's theorem - Introductory calculus: symbolic integration, arclength calculations - Vector calculus: gradients, divergence, curl - Multivariable calculus: Jacobians, Laplacians - Linear algebra: eigenvalues/eigenvectors, RREF, characteristic polynomials

      Using !!!ONLY!!! the given math concepts in the form, !!!DO NOT!!! include any guessings or conditions other than the concepts in the form!!!
      !!!Do NOT!!! propose conditions or special particularities. For example, a hypothesis like "The LLM is likely to fail on a concept, especially/particularly when the problem is...(some special conditions)" or "The LLM is likely to fail on a concept because of (some condition)" is not allowed!
      For example: "The LLM is likely to fail on problems that require function evaluation and transformations, especially when dealing with composite functions and inverses." is a BAD hypothesis that doesn't follow the previous requirement.
      Another example: "The LLM is likely to fail on (some math concept) when the problem is complex" is also a BAD hypothesis that doesn't follow the instructions. 'complex' is very vague and is a condition that is NOT in the math concept. It will NOT be accepted.
      !!!Do NOT!!! propose hypotheses only from a particular step in the problem! For example, a hypothesis like "The LLM is likely to fail on a concept at a (some step in problem solution)" is not allowed!
      !!!Do NOT!!! propose any reasons behind the failures! For example, a hypothesis like "The LLM is likely to fail on a concept, because/due to ... (some reason)" is not allowed!
      Again, Use !!!Only!!! the given math concepts in the form! Only means the hypotheses can only contain the concepts in the form! and no other things allowed!
      Using anything other than the math concepts in the form will NOT be accepted and should NOT be proposed!
      These hypotheses should identify specific patterns that occur across the provided problems and LLM-generated answers.
      please propose ${num_hypotheses} possible hypotheses.
      These hypotheses should identify specific patterns that occur across the provided problems and LLM-generated answers.
      You should check carefully the specific solving steps by the LLM, and consider which particular
      step and which particular math concept/skill did the LLM make mistake on.
      When proposing hypotheses, generate half hypotheses using a single math concept, and generate the other half by combining two or more math concepts.
      Again, use ONLY and VERBATIMLY the provided math concepts from the list

      Each hypothesis should be:
        A hypothesis about what particular math makes the LLM to fail.

      Generate them in the format of 1. [hypothesis], 2. [hypothesis], ... ${num_hypotheses}. [hypothesis].
      The hypotheses should analyze what particular math concept(s) are associated with correctness or error.

    user: |-
      We have seen some math problems and LLM-generated answers:
      ${observations}
      Please generate hypotheses that are useful for predicting which particular math concept and solution step does the LLM likely to make mistakes on.
      Propose ${num_hypotheses} possible hypotheses. Generate them in the format of 1. [hypothesis], 2. [hypothesis], ... ${num_hypotheses}. [hypothesis].
      Proposed hypotheses:

  inference:
    system: |-
      You are a professional math teacher and your job is to determine whether a given answer to a math problem is correct or wrong.
      From past experience, you have learned that LLMs are more likely to fail on certain math concepts (or combination of math concepts).
      You need to determine whether the learned pattern applies to the current problem and answer, and then make your prediction.
      Give your final answer in the format of "Final answer: answer", where the answer is either "correct" or "wrong".

    user: |-
      Our learned pattern: ${hypothesis}
      A math problem and its answer are the following:
      Problem: "${problems}"
      Answer: "${answers}"

      Given the pattern you learned above, decide whether the answer is correct or wrong.
      Think step by step.
      First step: Consider if the pattern can be applied to the answer.
      Second step: Based on the pattern, is this answer correct or wrong?
      Final step: give your final answer in the format of "Final answer: answer"

  multiple_hypotheses_inference:
    system: |-
      You are a professional math teacher and your job is to determine whether a given answer to a math problem is correct or wrong.
      From past experience, you have learned that LLMs are more likely to fail on certain math concepts (or combination of math concepts).
      You need to determine whether these patterns apply to the current problem and answer, and then make your prediction.
      Give your final answer in the format of "Final answer: answer", where the answer is either "correct" or "wrong".

    user: |-
      Our learned patterns: ${hypotheses}
      A math problem and its answer are the following:
      Problem: "${problems}"
      Answer: "${answers}"

      Given the patterns you learned above, decide whether the answer is correct or wrong.
      Think step by step.
      First step: Think about which pattern(s) can be applied to the answer.
      Second step: Based on the patterns, is this answer correct or wrong?
      Final step: give your final answer in the format of "Final answer: answer"
\end{minted}
\section{Granularity and Performance}
\label{appendix:granularity_performance}
\subsection{GPT4Omini}
\begin{figure}[H]
  \centering
  \begin{subfigure}[b]{0.45\textwidth}
    \includegraphics[width=\linewidth]{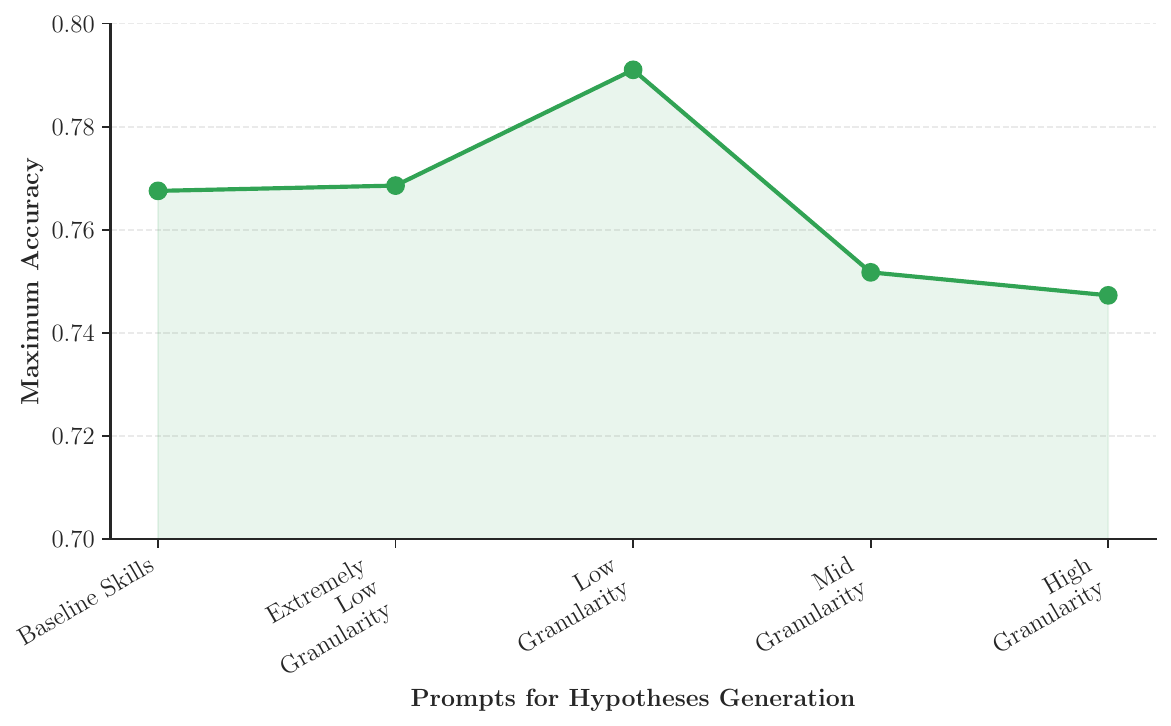}
    \caption{Maximum Accuracy Trend Using GPT4Omini}
  \end{subfigure}
  \hfill
  \begin{subfigure}[b]{0.45\textwidth}
    \includegraphics[width=\linewidth]{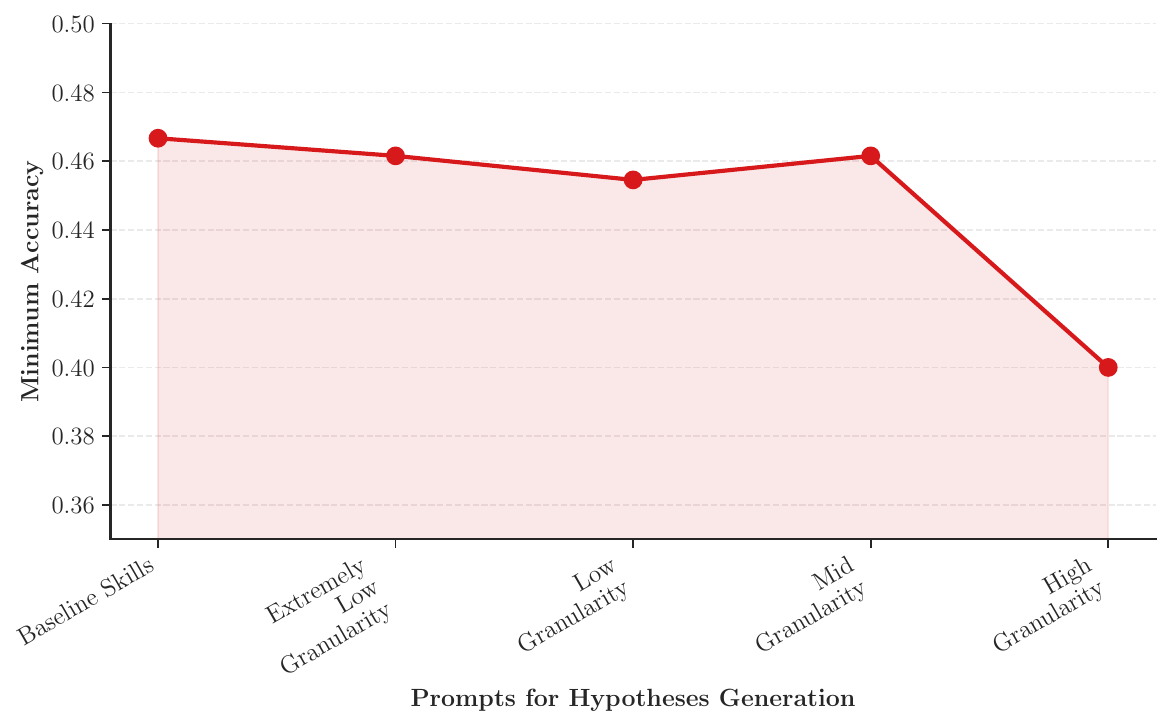}
    \caption{Minimum Accuracy Trend Using GPT4Omini}
  \end{subfigure}
  \caption{Hypotheses Accuracy Trending for GPT4Omini Model under Different Prompts for Hypotheses Generation}
\end{figure}
\begin{figure}[H]
  \centering
  \begin{subfigure}[b]{0.45\textwidth}
    \includegraphics[width=\linewidth]{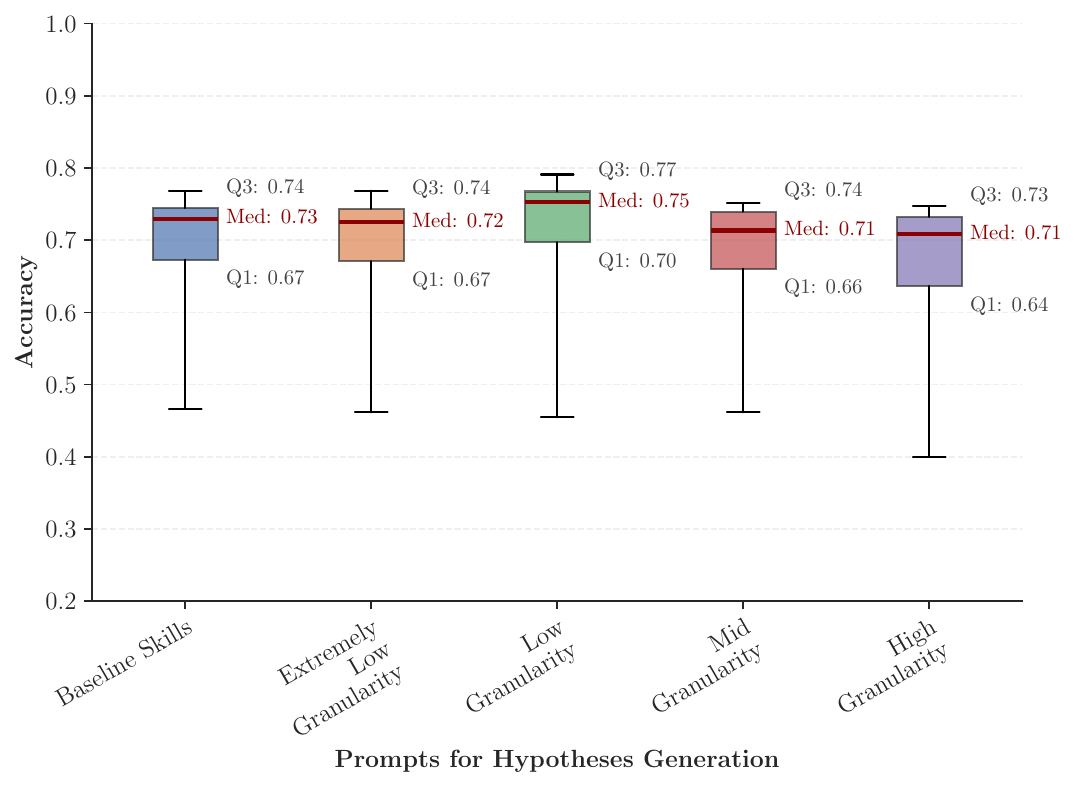}
    \caption{Boxplot for Hypotheses Accuracy Distribution Using GPT4Omini}
  \end{subfigure}
  \hfill
  \begin{subfigure}[b]{0.45\textwidth}
    \includegraphics[width=\linewidth]{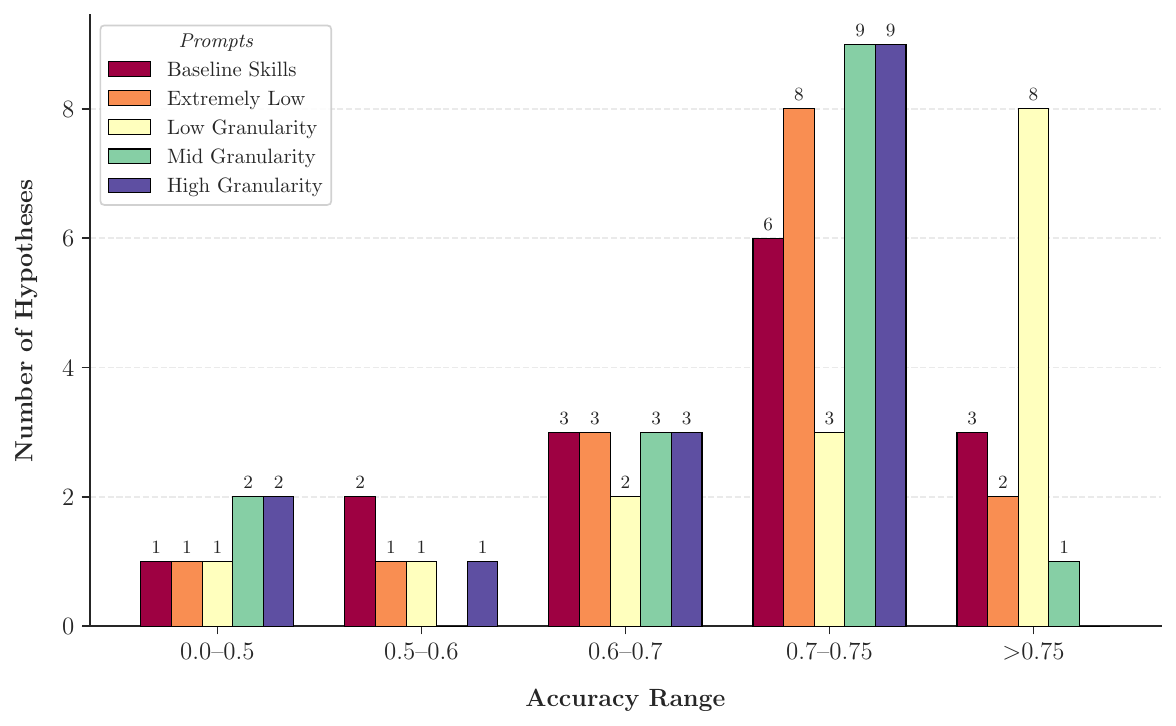}
    \caption{Counting of Hypotheses in Different Accuracy Range Using GPT4Omini}
  \end{subfigure}
  \caption{Distribution and Counting of Hypotheses Accuracies Using GPT4Omini under Different Prompts for Hypotheses Generation}
\end{figure}
\begin{figure}[H]
    \centering
    \includegraphics[width=0.8\linewidth]{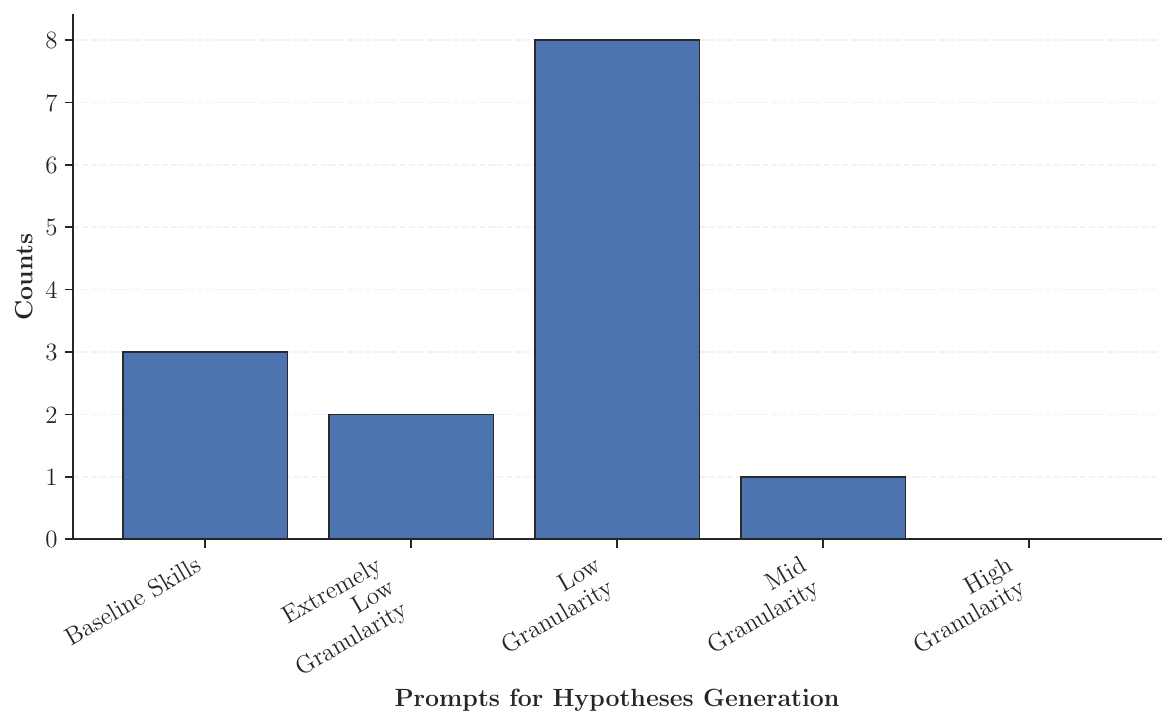}
    \caption{Number of Hypotheses with Accuracies Over 0.75 Using GPT4Omini under Different Prompts for Hypotheses Generation}
\end{figure}

\subsection{GPT4.1mini}
\label{subappendix:gpt4.1mini hypotheses distribution}
\begin{figure}[H]
  \centering
  \begin{subfigure}[b]{0.45\textwidth}
    \includegraphics[width=\linewidth]{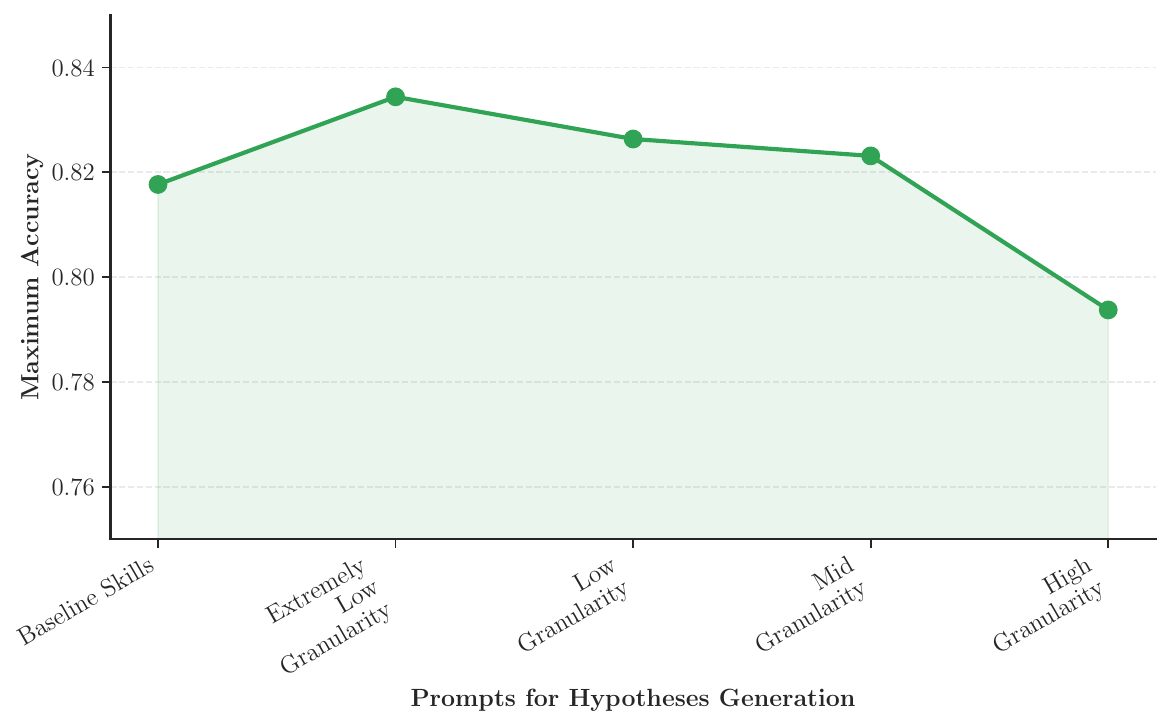}
    \caption{Maximum Accuracy Trend Using GPT4.1mini}
  \end{subfigure}
  \hfill
  \begin{subfigure}[b]{0.45\textwidth}
    \includegraphics[width=\linewidth]{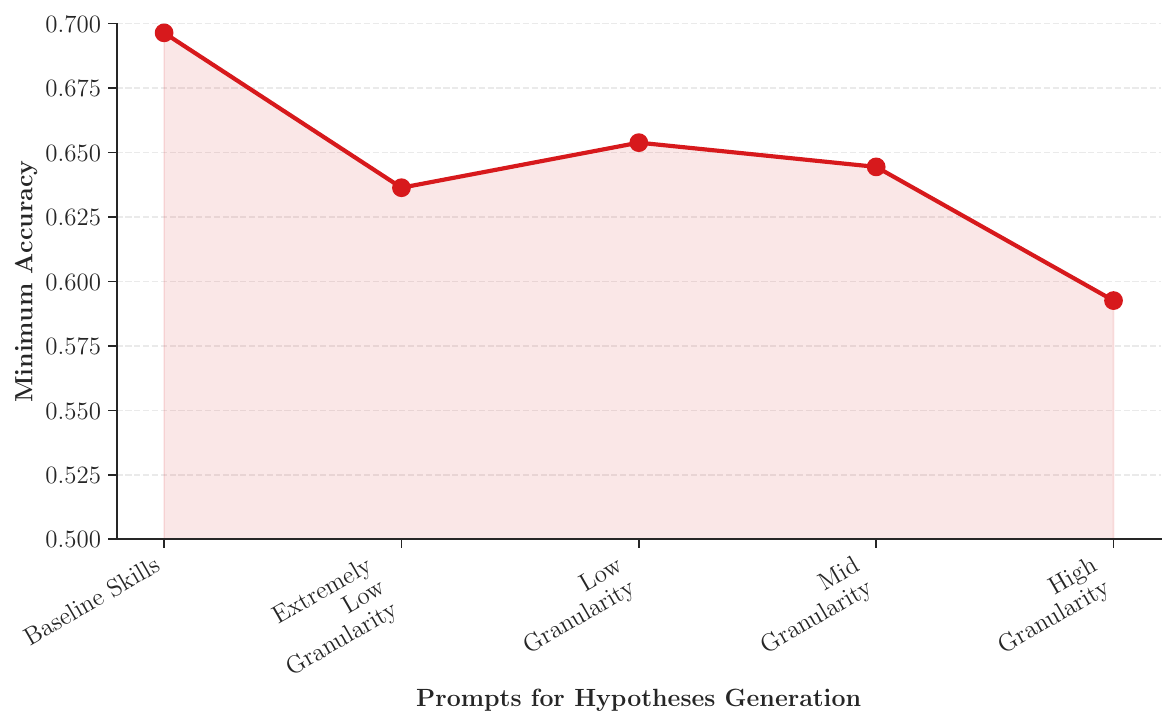}
    \caption{Minimum Accuracy Trend Using GPT4.1mini}
  \end{subfigure}
  \caption{Hypotheses Accuracy Trending for GPT4.1mini Model under Different Prompts for Hypotheses Generation}
\end{figure}
\begin{figure}[H]
  \centering
  \begin{subfigure}[b]{0.45\textwidth}
    \includegraphics[width=\linewidth]{Graphs/accuracy_boxplots_side_labels_4.1mini.pdf}
    \caption{Boxplot for Hypotheses Accuracy Distribution Using GPT4.1mini}
  \end{subfigure}
  \hfill
  \begin{subfigure}[b]{0.45\textwidth}
    \includegraphics[width=\linewidth]{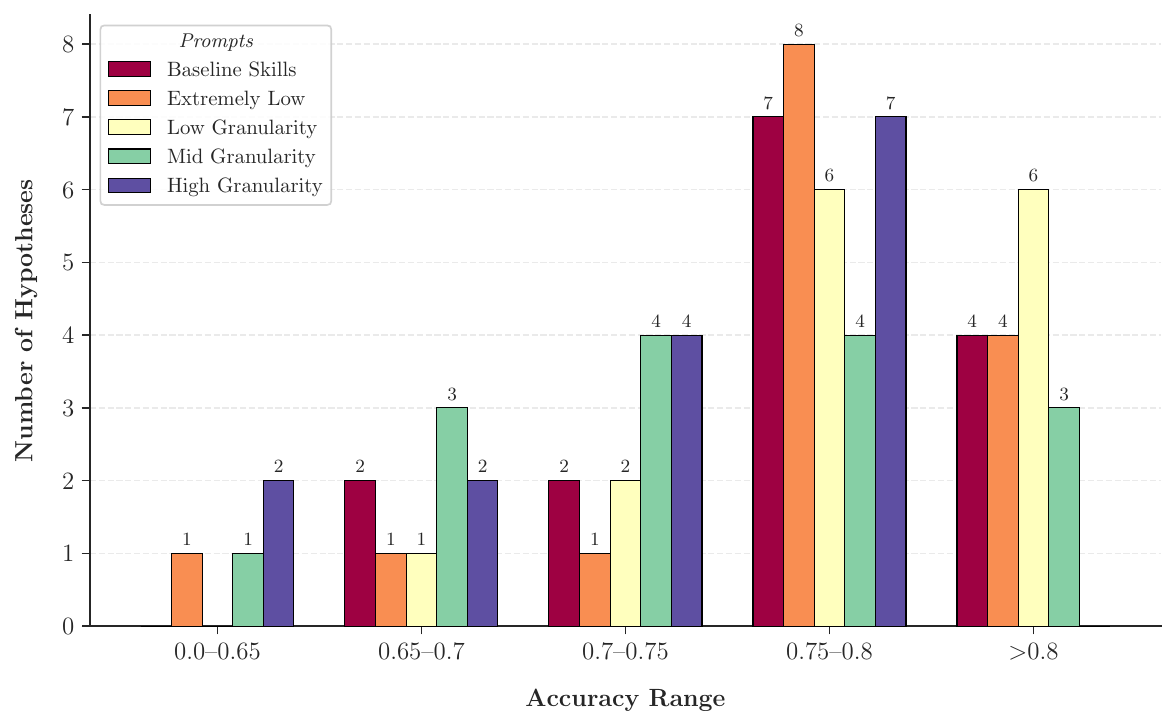}
    \caption{Counting of Hypotheses in Different Accuracy Range Using GPT4.1mini}
  \end{subfigure}
  \caption{Distribution and Counting of Hypotheses Accuracies Using GPT4.1mini under Different Prompts for Hypotheses Generation}
\end{figure}
\begin{figure}[H]
    \centering
    \includegraphics[width=0.8\linewidth]{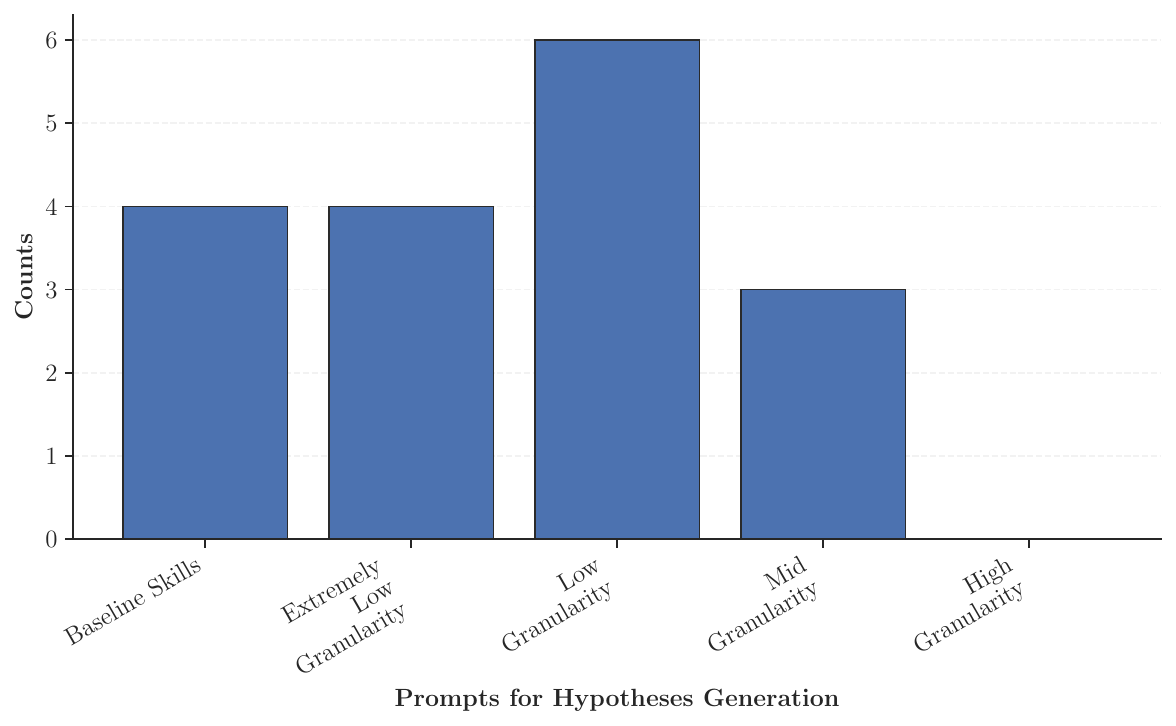}
    \caption{Number of Hypotheses with Accuracies Over 0.8 Using GPT4.1mini under Different Prompts for Hypotheses Generation}
    \label{fig:count of hypo>0.8 gpt4.1mini}
\end{figure}

\subsection{Qwen3-14B(Nonthinking mode)}
\begin{figure}[H]
  \centering
  \begin{subfigure}[b]{0.45\textwidth}
    \includegraphics[width=\linewidth]{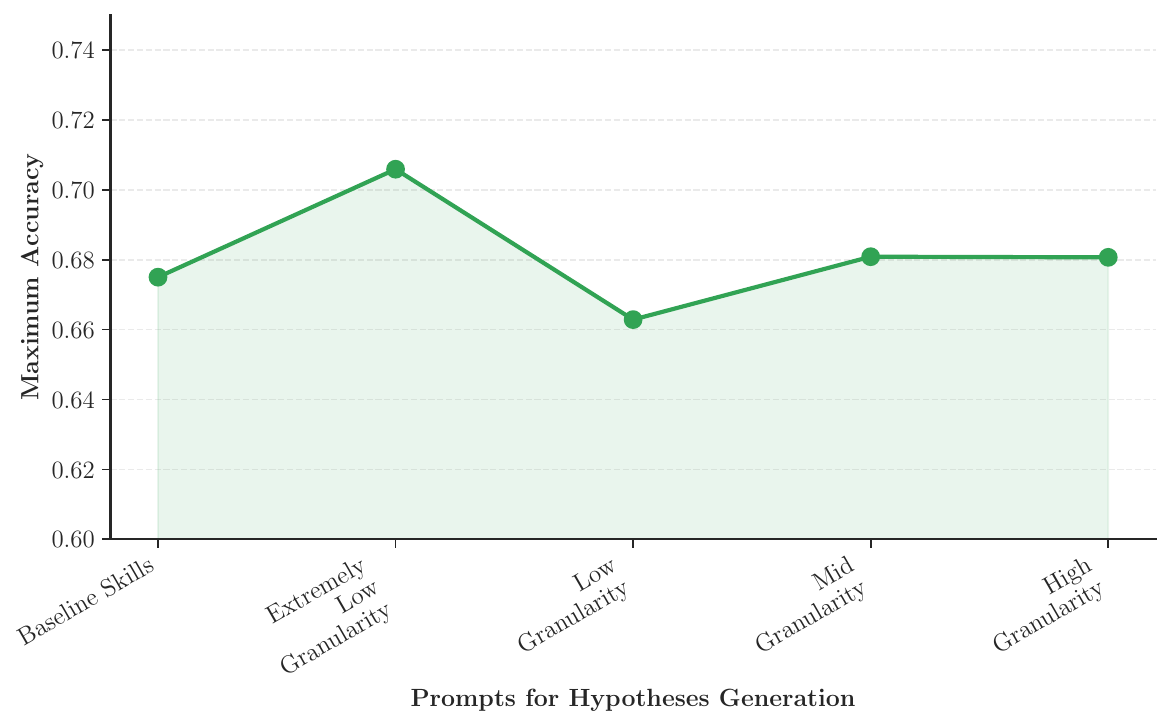}
    \caption{Maximum Accuracy Trend Using Qwen3-14B}
  \end{subfigure}
  \hfill
  \begin{subfigure}[b]{0.45\textwidth}
    \includegraphics[width=\linewidth]{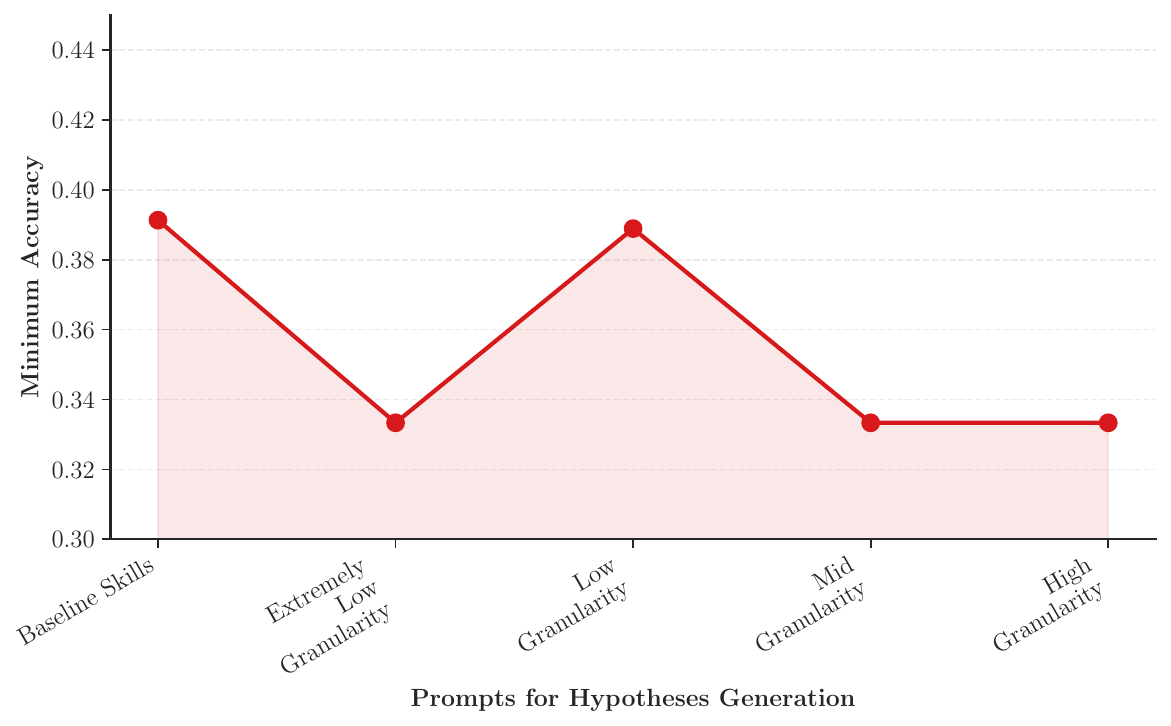}
    \caption{Minimum Accuracy Trend Using Qwen3-14B}
  \end{subfigure}
  \caption{Hypotheses Accuracy Trending for Qwen3-14B Model under Different Prompts for Hypotheses Generation}
\end{figure}
\begin{figure}[H]
  \centering
  \begin{subfigure}[b]{0.45\textwidth}
    \includegraphics[width=\linewidth]{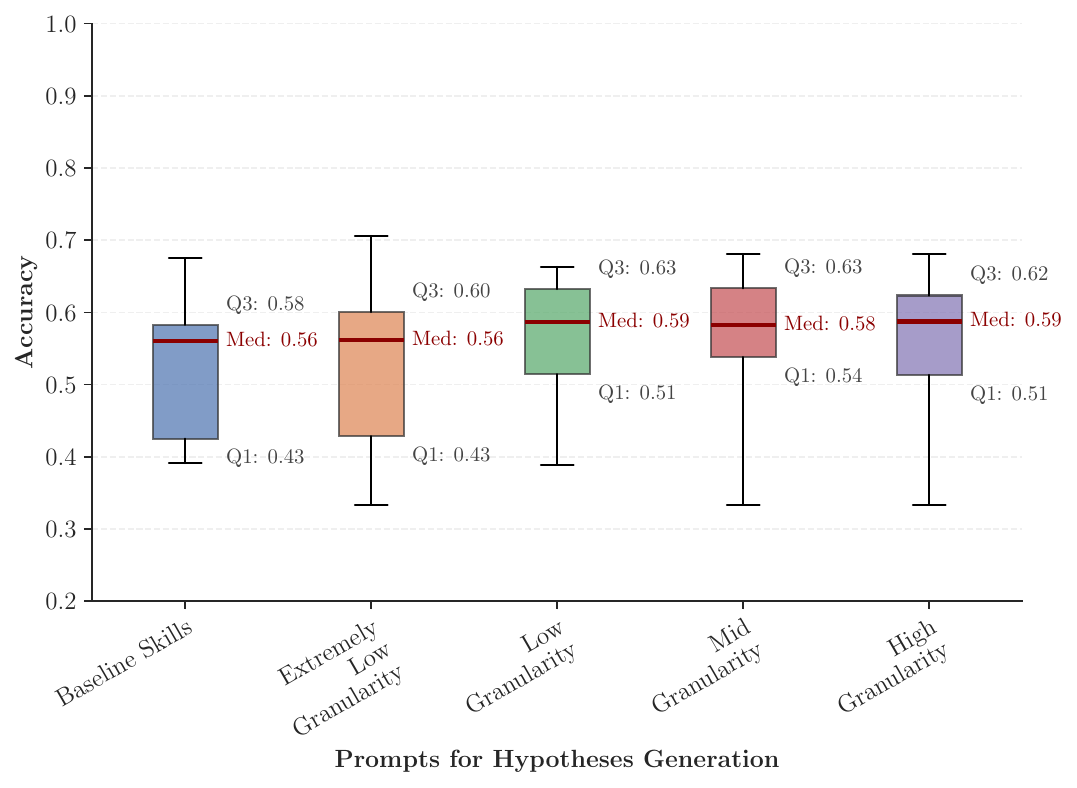}
    \caption{Boxplot for Hypotheses Accuracy Distribution Using Qwen3-14B}
  \end{subfigure}
  \hfill
  \begin{subfigure}[b]{0.45\textwidth}
    \includegraphics[width=\linewidth]{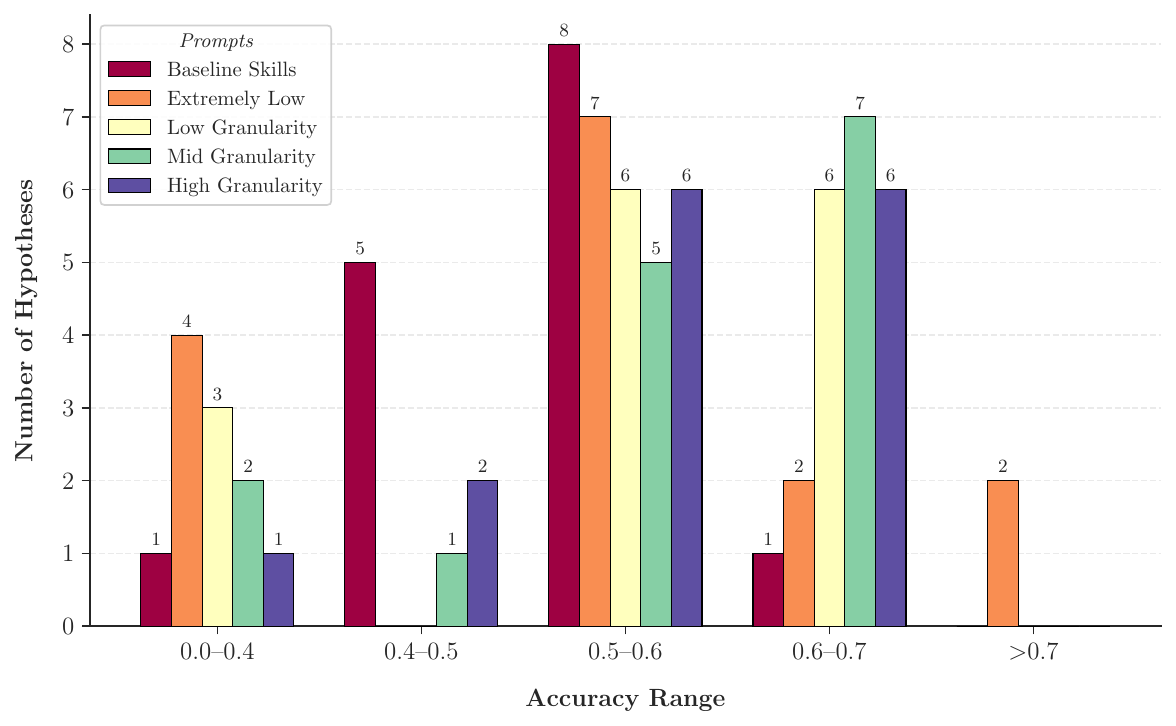}
    \caption{Counting of Hypotheses in Different Accuracy Range Using Qwen3-14B}
  \end{subfigure}
  \caption{Distribution and Counting of Hypotheses Accuracies Using Qwen3-14B under Different Prompts for Hypotheses Generation}
\end{figure}
\begin{figure}[H]
    \centering
    \includegraphics[width=0.8\linewidth]{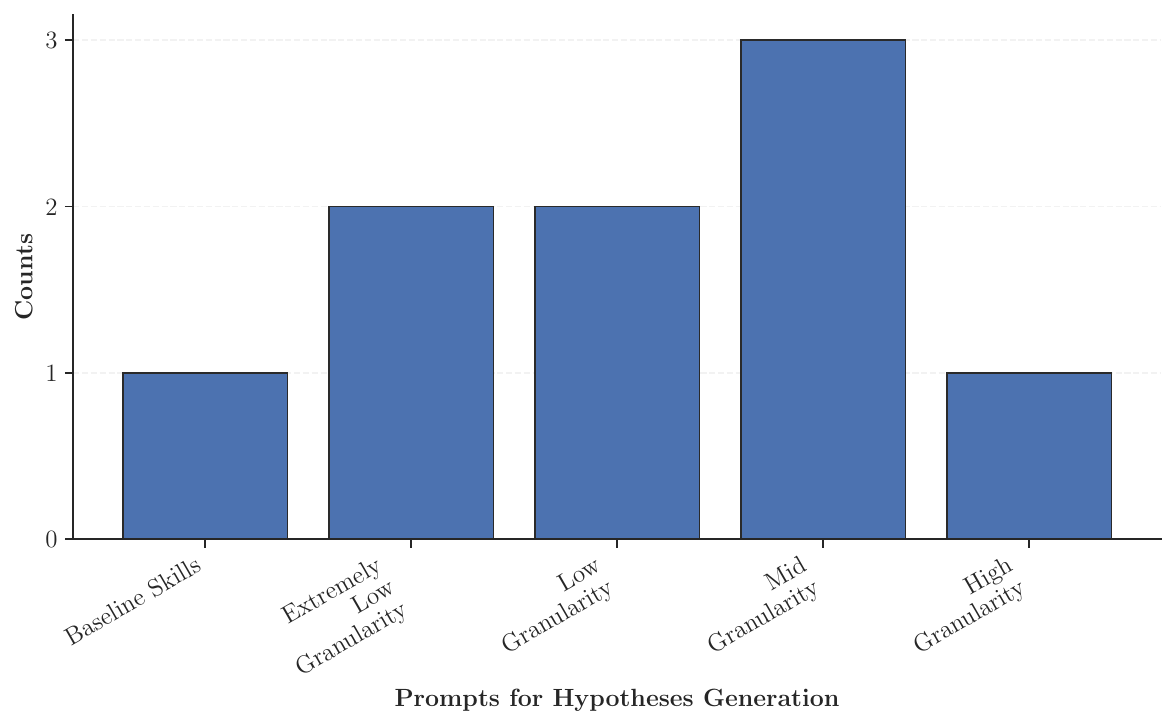}
    \caption{Number of Hypotheses with Accuracies Over 0.65 Using Qwen3-14B under Different Prompts for Hypotheses Generation}
\end{figure}

\section{Sample Generated Hypotheses}
\label{appendix:sample_hypotheses}
\subsection{Baseline Skills Hypotheses}
\label{subappendix:Baseline Skills Hypotheses}
\begin{minted}[fontsize=\normalsize, breaklines, breakanywhere]{yaml}
Hypothesis 1: "The LLM is likely to fail on problems involving combinatorics and probability skills combined with calculation and conversion skills." (accuracy: 0.6964285714285714)
Hypothesis 2: "The LLM is likely to fail on problems requiring calculation and conversion skills." (accuracy: 0.8176795580110499)
Hypothesis 3: "The LLM tends to fail on problems involving quadratic equations and solutions." (accuracy: 0.6981132075471698)
Hypothesis 4: "The LLM is likely to fail on problems that involve function composition and transformation combined with understanding and application of functions." (accuracy: 0.7900552486187845)
Hypothesis 5: "The LLM is likely to fail on problems involving solving system of equations." (accuracy: 0.7814569536423842)
Hypothesis 6: "The LLM is likely to fail on problems involving geometry triangle properties combined with trigonometry skills." (accuracy: 0.7888198757763973)
Hypothesis 7: "The LLM is likely to fail on problems involving number theory and arithmetic operations." (accuracy: 0.797546012269939)
Hypothesis 8: "The LLM is likely to fail on problems involving geometric relations and triangle geometry skills." (accuracy: 0.7642857142857142)
Hypothesis 9: "The LLM is likely to fail on problems combining algebraic manipulation skills and solving equations." (accuracy: 0.8022598870056498)
Hypothesis 10: "The LLM is likely to fail on problems requiring solving system of equations." (accuracy: 0.8048780487804879)
Hypothesis 11: "The LLM is likely to fail on problems involving arithmetic sequences combined with sequence and series skills." (accuracy: 0.7341772151898734)
Hypothesis 12: "The LLM shows difficulty on problems involving coordinate geometry and transformation skills together with graph understanding and interpretation." (accuracy: 0.8157894736842105)
Hypothesis 13: "The LLM is likely to fail on problems involving combinatorics and probability skills." (accuracy: 0.77)
Hypothesis 14: "The LLM is likely to fail on problems involving probability concepts and calculations and permutation and combinations." (accuracy: 0.77)
Hypothesis 15: "The LLM is likely to fail on problems involving solving system of equations and algebraic expression skills." (accuracy: 0.7090909090909091)
\end{minted}

\subsection{Extremely Low Granularity Hypotheses}
\label{subappendix:Extremely Low Granularity Hypotheses}
\begin{minted}[fontsize=\normalsize, breaklines, breakanywhere]{yaml}
Hypothesis 1: "The LLM is likely to fail on problems involving the combination of Geometry and Algebra." (accuracy: 0.8343949044585988)
Hypothesis 2: "The LLM is likely to fail on problems involving the combination of Number Theory, Geometry, and Counting & Probability." (accuracy: 0.7737226277372263)
Hypothesis 3: "The LLM is likely to fail on problems involving the combination of Algebra, Number Theory, and Counting & Probability." (accuracy: 0.7737226277372263)
Hypothesis 4: "The LLM is likely to fail on problems involving the combination of Prealgebra and Counting & Probability." (accuracy: 0.7870370370370369)
Hypothesis 5: "The LLM is likely to fail on problems involving a combination of Counting & Probability and Geometry." (accuracy: 0.7788461538461539)
Hypothesis 6: "The LLM is likely to fail on problems involving a combination of Number Theory and Precalculus." (accuracy: 0.776)
Hypothesis 7: "The LLM is likely to fail on problems involving both Geometry and Precalculus." (accuracy: 0.7671232876712328)
Hypothesis 8: "The LLM is likely to fail on problems involving both Prealgebra and Algebra." (accuracy: 0.8295454545454546)
Hypothesis 9: "The LLM is likely to fail on problems involving a combination of Geometry and Precalculus." (accuracy: 0.7985611510791367)
Hypothesis 10: "The LLM is likely to fail on problems involving Prealgebra." (accuracy: 0.6904761904761905)
Hypothesis 11: "The LLM is likely to fail on problems involving the combination of Geometry and Counting & Probability." (accuracy: 0.7368421052631579)
Hypothesis 12: "The LLM is likely to fail on problems involving the combination of Prealgebra, Algebra, and Geometry." (accuracy: 0.7972972972972971)
Hypothesis 13: "The LLM is likely to fail on problems involving Number Theory, Geometry, and Precalculus." (accuracy: 0.8064516129032258)
Hypothesis 14: "The LLM is likely to fail on problems involving a combination of Prealgebra and Algebra." (accuracy: 0.6363636363636364)
Hypothesis 15: "The LLM is likely to fail on problems involving the combination of Prealgebra and Algebra." (accuracy: 0.8071428571428572)
\end{minted}

\subsection{Low Granularity Hypotheses}
\label{subappendix:Low Granularity Hypotheses}
\begin{minted}[fontsize=\normalsize, breaklines, breakanywhere]{yaml}
Hypothesis 1: "The LLM is likely to fail on problems involving Spatial reasoning, geometric theorem application." (accuracy: 0.8092105263157894)
Hypothesis 2: "The LLM is more likely to make mistakes on problems involving Combinatorics, probability modeling." (accuracy: 0.6538461538461539)
Hypothesis 3: "The LLM is likely to fail on problems involving Multi-step reasoning, deduction, diagram use." (accuracy: 0.7851239669421488)
Hypothesis 4: "The LLM is likely to fail on problems involving multi-step reasoning, deduction, diagram use." (accuracy: 0.8156424581005588)
Hypothesis 5: "The LLM is likely to fail on problems involving Modular arithmetic, divisibility, integer properties." (accuracy: 0.8226950354609929)
Hypothesis 6: "The LLM is likely to fail on problems that require Combinatorics, probability modeling and Spatial reasoning, theorem application." (accuracy: 0.7272727272727273)
Hypothesis 7: "The LLM is likely to fail on problems involving spatial reasoning, geometric theorem application." (accuracy: 0.8225806451612904)
Hypothesis 8: "The LLM is likely to fail on problems that require spatial reasoning, theorem application combined with sequences, function analysis." (accuracy: 0.8206521739130435)
Hypothesis 9: "The LLM is likely to fail on problems that require Multi-step reasoning, deduction, diagram use and parabolas, ellipses, hyperbolas, GCD." (accuracy: 0.7904761904761904)
Hypothesis 10: "The LLM is likely to fail on problems that require both Expression manipulation, equation solving and Sequences, function analysis." (accuracy: 0.782178217821782)
Hypothesis 11: "The LLM is likely to fail on problems involving Combinatorics, probability modeling." (accuracy: 0.7798165137614679)
Hypothesis 12: "The LLM is likely to fail on problems involving multi-step reasoning, deduction, diagram use combined with sequences, function analysis." (accuracy: 0.8263473053892217)
Hypothesis 13: "The LLM is more likely to make mistakes on problems involving both Sequences, function analysis and Expression manipulation, equation solving." (accuracy: 0.7978723404255318)
Hypothesis 14: "The LLM is likely to fail on problems involving Sequences, function analysis." (accuracy: 0.7777777777777778)
Hypothesis 15: "The LLM is likely to fail on problems that require both Modular arithmetic, divisibility, integer properties and Spatial reasoning, theorem application." (accuracy: 0.723404255319149)
\end{minted}

\subsection{Mid Granularity Hypotheses}
\label{subappendix:Mid Granularity Hypotheses}
\begin{minted}[fontsize=\normalsize, breaklines, breakanywhere]{yaml}
Hypothesis 1: "The LLM is more error-prone on problems involving function eval/composition." (accuracy: 0.8231292517006804)
Hypothesis 2: "The LLM is more likely to fail on problems requiring the use of linear & systems concepts." (accuracy: 0.8200000000000002)
Hypothesis 3: "The LLM is more error-prone on problems that involve angles and triangle theorems." (accuracy: 0.8195876288659791)
Hypothesis 4: "The LLM is more likely to make errors on problems involving polynomial ops and factoring quadratics." (accuracy: 0.7999999999999998)
Hypothesis 5: "The LLM is more likely to make mistakes on problems involving Linear & systems." (accuracy: 0.7877094972067039)
Hypothesis 6: "The LLM shows increased likelihood of error when problems involve sequences & series." (accuracy: 0.7772020725388601)
Hypothesis 7: "The LLM tends to make mistakes on problems involving linear & systems." (accuracy: 0.75)
Hypothesis 8: "The LLM is more likely to make mistakes on problems combining exponential/logarithmic equations and function inversion/transformation." (accuracy: 0.6941176470588237)
Hypothesis 9: "The LLM is more likely to make mistakes on problems involving Linear & systems together with inequalities." (accuracy: 0.7222222222222223)
Hypothesis 10: "The LLM is more likely to fail on problems involving polynomial division." (accuracy: 0.6444444444444443)
Hypothesis 11: "The LLM is more likely to make mistakes on problems involving polynomial division and factor theorem." (accuracy: 0.7102803738317754)
Hypothesis 12: "The LLM is more likely to make mistakes on problems involving function evaluation/composition combined with function inversion/transformation." (accuracy: 0.7523809523809525)
Hypothesis 13: "The LLM is more likely to produce incorrect answers on problems involving GCD and LCM." (accuracy: 0.676923076923077)
Hypothesis 14: "The LLM is more error-prone on problems involving circle theorems." (accuracy: 0.6874999999999999)
Hypothesis 15: "The LLM tends to fail on problems involving sequences & series." (accuracy: 0.7272727272727274)
\end{minted}

\subsection{High Granularity Hypotheses}
\label{subappendix:High Granularity Hypotheses}
\begin{minted}[fontsize=\normalsize, breaklines, breakanywhere]{yaml}
Hypothesis 1: "The LLM is likely to fail on problems combining Integer arithmetic (+,-,*,/) and Order of operations (PEMDAS)." (accuracy: 0.7837837837837838)
Hypothesis 2: "The LLM is likely to fail on problems involving Logarithmic functions, properties, and log rules." (accuracy: 0.6666666666666666)
Hypothesis 3: "The LLM is likely to fail on problems involving Function evaluation and basic transformations." (accuracy: 0.7937499999999996)
Hypothesis 4: "The LLM is likely to fail on problems involving Integer arithmetic (+,-,*,/)." (accuracy: 0.7875000000000001)
Hypothesis 5: "The LLM is likely to fail on problems involving Factorials (n!) and Permutations P(n,k)." (accuracy: 0.7720588235294118)
Hypothesis 6: "The LLM is likely to fail on problems involving Absolute value equations and inequalities." (accuracy: 0.7456140350877193)
Hypothesis 7: "The LLM is likely to fail on problems involving systems of linear equations solved by substitution or elimination." (accuracy: 0.761111111111111)
Hypothesis 8: "The LLM is likely to fail on problems that combine Integer exponents (e.g., 2^n, 3^n) and Square roots & cube roots (perfect and approximations)." (accuracy: 0.7517241379310344)
Hypothesis 9: "The LLM is likely to fail on problems involving systems of linear equations (substitution/elimination)." (accuracy: 0.7560975609756095)
Hypothesis 10: "The LLM is likely to fail on problems combining Complex algebraic manipulations of expressions/inequalities with Quadratic & higher-degree inequalities (sign analysis, test intervals)." (accuracy: 0.7372881355932204)
Hypothesis 11: "The LLM is likely to fail on problems involving polynomial long division and the factor theorem or remainder theorem." (accuracy: 0.7342657342657342)
Hypothesis 12: "The LLM is likely to fail on problems involving both Quadratic solving and Systems mixing linear and quadratic equations." (accuracy: 0.7073170731707317)
Hypothesis 13: "The LLM is likely to fail on problems requiring multi-step linear equations." (accuracy: 0.5925925925925926)
Hypothesis 14: "The LLM is likely to fail on problems combining Prime identification & prime factorization with Divisibility tests (2-11)." (accuracy: 0.6734693877551019)
Hypothesis 15: "The LLM is likely to fail on problems combining Conic sections: properties of parabolas, ellipses, hyperbolas and Analytical geometry: shifts, intersections, slopes in coordinate plane." (accuracy: 0.6363636363636364)
\end{minted}

\section{Problem Generation Prompts}
\label{appendix:problem_generation_prompts}
\begin{minted}[fontsize=\normalsize, breaklines, breakanywhere]{yaml}
Your task is to make some math problems to evaluate students' performance on certain math concepts and skills.
You decide to make the problems more challenging, so that you can easily test students' understanding
of these math concepts/skills. You have also found certain rules that would make the problems more difficult and challenging.
You found that students are very likely to fail on:
    <one Hypothesis>
Therefore, you decide to make geometric spatial reasoning problems that involves the application of different geometric theorem so that the problems would be challenging and fully reveal the students' abilities.
You decide to make a total of {n} such problems that are similar to problems in the MATH benchmark.
Each problem should:
    1. Be clear and non-ambiguous.
    2. Be correct and have no factual mistakes. Check carefully before giving the final questions.
    3. Express clearly so that the students can understand what you mean.
    4. The problems should have a specific answer, instead of being a proof question. NO PROOF QUESTIONS!
    5. No true and false problems, yes or no problems, or explanation problems. The problem should have a specific numeric or binary answer.
    6. Make the problems challenging instead of just doing very simple arithmetic. You can try to involve multiple steps or math concepts/skills in a single problem.
When giving the problem statements, you MUST state them in the format: "Problem 1/2/3/...:". Each problem should be separated by a blank line. You must make exactly {n} questions. No more, no less."""
\end{minted}

\section{Sample Generated Math Problems}
\label{appendix:sample_generated_math_problems}
\subsection{Baseline Skill Granularity Generated Problems}
Hypothesis used: ``The LLM is likely to fail on problems requiring calculation and conversion skills.''
\begin{minted}[fontsize=\normalsize, breaklines, breakanywhere]{yaml}
"A car travels from City A to City B at an average speed of 60 km/h and returns at an average speed of 40 km/h. What is the average speed of the car for the entire trip if the distance between City A and City B is 240 km?",
"A bakery sells 250 loaves of bread per day. Each loaf of bread requires 0.5 kg of flour. If the price of flour is $1.50 per kg, how much does the bakery spend on flour per day?",
"A person invests $5000 in a savings account that earns an annual interest rate of 4%
"A rectangular prism has a length of 8 cm, a width of 5 cm, and a height of 3 cm. What is the volume of the prism in cubic centimeters?",
"A water tank can hold 1200 liters of water. If 3/4 of the tank is already filled, how many more liters of water can be added to fill the tank?",
"A group of friends want to share some candy equally. If they have 48 pieces of candy and there are 8 friends, how many pieces of candy will each friend get?",
"A person has $1000 to invest in two different investments. Investment A earns an annual interest rate of 5%
"A cylinder has a radius of 4 cm and a height of 10 cm. What is the volume of the cylinder in cubic centimeters?",
"A person buys 15 meters of fabric at $8 per meter. If the person also buys 20 buttons at $0.50 each, what is the total cost of the fabric and buttons?",
"A car travels 250 miles in 5 hours. If the car uses 25 gallons of gas, how many miles per gallon does the car get?",
"A bakery sells 300 loaves of bread per day. Each loaf of bread requires 0.75 kg of flour. If the price of flour is $2.00 per kg, how much does the bakery spend on flour per day?",
"A person invests $2000 in a savings account that earns an annual interest rate of 6%
"A rectangular prism has a length of 10 cm, a width of 6 cm, and a height of 4 cm. What is the volume of the prism in cubic centimeters?",
"A water tank can hold 1500 liters of water. If 2/3 of the tank is already filled, how many more liters of water can be added to fill the tank?",
"A group of friends want to share some candy equally. If they have 72 pieces of candy and there are 12 friends, how many pieces of candy will each friend get?",
"A person has $2000 to invest in two different investments. Investment A earns an annual interest rate of 4%
"A sphere has a radius of 6 cm. What is the volume of the sphere in cubic centimeters?",
"A person buys 20 meters of fabric at $10 per meter. If the person also buys 30 buttons at $0.75 each, what is the total cost of the fabric and buttons? (to 2 decimal places)",
"A car travels 300 miles in 6 hours. If the car uses 30 gallons of gas, how many miles per gallon does the car get?",
"A bakery sells 400 loaves of bread per day. Each loaf of bread requires 0.5 kg of flour. If the price of flour is $2.50 per kg, how much does the bakery spend on flour per day?"
\end{minted}

\subsection{Extremely Low Granularity Generated Problems}
Hypothesis used: ``LLMs are likely to fail on problems involving the combination of Geometry and Algebra.''
\begin{minted}[fontsize=\normalsize, breaklines, breakanywhere]{yaml}
"In a right triangle with legs measuring 8cm and 15cm, find the length of the hypotenuse using algebraic expression.",
"Find the area of an equilateral triangle inscribed within a circle of radius 12cm, where each side of the triangle touches the circumference of the circle.",
"Determine the equation for a line passing through two points (x1,y1) = (-3,-2) and (x2,y2) = (4,5), and use it to calculate the x-coordinate at which this line intersects a horizontal line y=0.",
"Calculate the volume of a pyramid whose base is a square with sides of length 10 cm and whose height is given as h = -b^2 + 30b -100, where b represents the length of one side of the square base.",
"Use geometry to solve for the value of x (rounded to the nearest degree) in the equation sin(x) = sqrt(1-cos^2(x)), given that cos(x) = 3/5.",
"A cube has side lengths of s cm, and its surface area decreases by 16%
"Derive the vertex of a parabola given its focus F=(h,k)=(2,3) and directrix line d:x=-5.",
"Solve for z in the parametric equations representing the intersection of a plane and a line. Plane P is represented as ax+by+cz+d=0; Line L passes through point Q=(q,r,s)=(1,2,-1); and direction vector n=<n_x,n_y,n_z> = <2,1,-3>. Assume coefficients: a=3,b=-2,c=1,d=9.",
" Calculate the arc length (to four decimal places) along a curve defined by r(t)=t^2+t for t in [a,b] where [a,b]=[0,pi], considering the Cartesian coordinates and the polar coordinates and give the sum of the two results.",
"Calculate the sum of areas of 25 congruent sectors cut out from a large circular disc of radius R=40cm. Given central angle theta=72 degrees corresponding to one sector.",
"Given parallel lines l1:x-y=2 and l2:x+y=6 intersecting a line l3:y=x+1, find the sum (to three decimal places) of their respective distances to the origin O(0,0).",
"If a cone is circumscribing a sphere, determine the sum of the radii r1, r2 if it is known that they share the same height H=10 and the ratio of volumes V_cone/V_sphere is equal to 27.",
"In the given diagram, two triangles ABC and ADE are shown. The coordinates of points A, B, C, D, and E are (0,0), (2,4), (4,2), (1,1), and (3,3) respectively. Find the area of triangle ADE.",
"With reference to Problem 7, substitute the value of k back into the original equation for calculating p. Then consider finding another point lying upon parabola by taking a suitable average value (for instance k=p/q etc.) from its standard form.",
"Evaluate the triple integral over region D defined as x^2+y^2+z^2<=R^2 of function f(x,y,z)=z/(x^2+y^2+z^2)^(3/2), knowing R=5 units.",
"Two lines intersect at a point P. The equations of the two lines are 2x + 3y = 7 and x - 2y = -3. Find the coordinates of point P.",
"A circle centered at C(0,h) intersects line y=h+sqrt(3x). Assuming r is the distance from center to (0,0), find expression relating r & h under specified constraints.",
"Calculate numerically the derivative dy/dx of implicit function: 3xy+x^3-y^2=7 for point (x_0,y_0)=(2,1)",
"Find the area of a trapezoid with bases of length 8 and 12 and height of 6. Find its area.",
"In parallelogram ABCD, the coordinates of points A, B and C are (1,1), (4,4), and (6,4) respectively. Find the coordinates of point D."
\end{minted}

\subsection{Low Granularity Generated Problems}
Hypothesis used: ``LLMs are likely to fail on problems involving Modular arithmetic, divisibility, integer properties.''
\begin{minted}[fontsize=\normalsize, breaklines, breakanywhere]{yaml}
"If $a \\equiv b \\pmod{m}$, then is it true that $(ab)^n \\equiv (ba)^n \\pmod{m}$ for any positive integers n?",
"Determine the remainder when $7^{100}$ is divided by 25.",
"Find all integers x satisfying $x^2 + 6x - 8 \\equiv 0 \\pmod{11}$.",
"What is the smallest positive integer N such that $N \\equiv 1 \\pmod{4}$, $N \\equiv 1 \\pmod{9}$, and $N \\equiv 2 \\pmod{13}$?",
"For how many integers $x$ where $1 \\leq x < 1000$, does $x^2 \\equiv 12 \\pmod{19}$ not hold true?",
"Let m = 13 * 7 * 3. How many solutions exist for $x^2 \\equiv 16 \\pmod{m}$?",
"Evaluate the expression $(10^{18} + 45) \\pmod{23}$.",
"Given $f(x) \\equiv ax + b \\pmod{29}$ and $g(x) \\equiv cx + d \\pmod{29}$, find the value of c given f(3) = g(3), a = 14, b = 27, d = 22.",
"For how many values of m does $36m + 27$ yield a number divisible by 9 but not 81, where m is an integer between 1 and 100 inclusive?",
"In the equation $3x^2 + 2y^2 \\equiv 15 \\pmod{23}$, determine the number of possible remainders of x when divided by 23.",
"We define a sequence {an} as follows: $a_n= 3a_{n-1}+ 4$. If a_0 = 1, find $a_{2019} \\pmod{7}$.",
"Find the number of distinct values for $a$ such that the congruence $x^2 \\equiv a \\pmod{101}$ has at least two distinct solutions.",
"Consider the system $x \\equiv 2 \\pmod{7}$, $x \\equiv 0 \\pmod{3}$, $x \\equiv 6 \\pmod{17}$. Find the unique solution $\\pmod{357}$.",
"Solve for x: $9x \\equiv 30 \\pmod{99}$",
"Consider the polynomial $f(x)= 4x^4 + ax^3 + bx^2 + cx + 21$, with integer coefficients a, b, c. If r is an integer root of this polynomial, find the number of possible remainders of r modulo 7.",
"Evaluate $(64^{201} + 37) \\pmod{97}$.",
"A positive integer n is called 'good' if the sum of its digits is a multiple of 7. Determine how many good numbers exist between 100000 and 999999, inclusive.",
"Determine the remainder when $7^{100} + (-8)^{200}$ is divided by 15.",
"How many four-digit multiples of 16 are there which do NOT contain the digit 5?",
"Given that p and q are prime numbers such that pq = 35, find the value of $p^q + q^p$ mod 12."
\end{minted}

\subsection{Mid Granularity Generated Problems}
Hypothesis used: ``LLMs are more error-prone on problems involving function evaluation/composition.''
\begin{minted}[fontsize=\normalsize, breaklines, breakanywhere]{yaml}
"A bakery sells a total of 250 loaves of bread per day. Among the breads, wheat bread has price 5 and white bread has price 7. If they sell a total of 1510, how many white breads did they sell?",
"Solve for y when 2x + 5y = 11 and x - 2y = -3.",
"Find the value of z in the system of equations: \nz + 2w = 7\nz - w = 1",
"In a factory, there are two types of machines, X and Y. Machine X produces 200 units per hour and machine Y produces 300 units per hour. The cost of running machine X is $100 per hour while the cost of running machine Y is $150 per hour. If the factory operates for 8 hours and wants to produce at least 2800 units with a maximum budget of $1200, how many hours should each type of machine run?",
"A car rental company has two types of cars, compact and SUV. The rental fee for a compact car is $40 per day and an SUV costs $60 per day. The company has 50 compact cars and 30 SUVs available. If the total revenue from renting out all the compact cars and 20 SUVs is $3400, how much money will the company make if it rents out 25 compact cars and all the SUVs?",
"Given two equations:\nx + 2y - 3z = 5\n2x - y + z = 3\nFind the value of x when y = 2 and z = 1.",
"Tom has been saving money for a new bike and has $120 in his piggy bank. He wants to buy a bike that costs $180. His parents have agreed to give him an allowance of $5 per week for doing chores. However, he spends $2 per week on snacks. How many weeks will it take Tom to save enough money to buy the bike?",
"A store has two types of discounts: 10%
"A group of friends want to split a bill evenly. If the total bill comes out to be $240 and there are 8 friends, but 2 of them didn't eat anything, how much will each friend who ate pay?",
"A farmer has 15 acres of land and plants two types of crops, corn and soybeans. Corn requires 20 gallons of water per acre and soybeans require 15 gallons per acre. If the farmer has 450 gallons of water available, how many acres of corn can he plant?",
"Solve for x in the equation:\n2x + 3(5 - x) = 14",
"If it takes 5 machines 5 minutes to make 5 widgets, how long will it take 100 machines to make 100 widgets?",
"A bakery is having a sale on bread. A loaf of bread normally costs $2, but during the sale, it's 15%
"Solve the system of equations:\nx + y + z = 6\n2x - y - z = 0\n-x + 2y - 3z = -2",
"A student scored 75, 85, and 90 on three quizzes. What score must the student get on the next quiz to have an average of 85?",
"In a right triangle, the length of the hypotenuse is 10 inches and one leg is 6 inches. What is the length of the other leg?",
"A company has two departments: sales and marketing. The sales department has 20 employees and the marketing department has 15 employees. The average salary in the sales department is $40000 and the average salary in the marketing department is $50000. What is the total payroll for the company?",
"Solve for y in the equation:\ny - 2(y - 3) = 7",
"A cyclist travels 20 miles in 2 hours. If she increases her speed by 25%
"Solve the equation:\n(x + 2)(x - 3) = 0 for x."
\end{minted}

\subsection{High Granularity Generated Problems}
Hypothesis used: ``The LLM is more likely to make mistakes on "problems involving function evaluation and basic transformations.''
\begin{minted}[fontsize=\normalsize, breaklines, breakanywhere]{yaml}
"Find f(-x) if f(x)= (2*x^2 + x)/(x+7)",
"Given g(t) = t^2 - 4t + 9 and h(x) = 3x - 12, find (g composite h)(0)",
"If m(a) = 8a - 11 and n(b) = b + 15, then m(n(2)) equals",
"Let p(u) = u^2 - 10u + 16 and q(v) = v - 14, find p(q(21))",
"Evaluate f(6) for f(x) = |x - 4| / (x - 3)",
"For r(w) = w^2 - 13w + 40 and s(y) = y^2 + 22y - 24, calculate the product (r*s)(-2)",
"Determine k(z) when z=5 if k(z) = z/(z-9) + 17/z",
"Solve j(c) where c=-3 if j(c) = sqrt((c^2)-25)/c+3, where j(c) can be complex.",
"If d(k) = k^3 - 6k^2 + 9k - 10 and e(j) = 5j + 18, find d(e(-2))",
"Calculate i(l) if l=2 and i(l)=(l+5)^2/l-5",
"Evaluate b(f) where f=10 for b(f) = abs(f-19)/f+7",
"Given o(p) = p^2 - 2p + 5 and t(d) = d^2 - 7d - 1, find the product (o*t)(5)",
"Let a(r) = 7*r^2 + 15*r + 29, evaluate a(2r) given r=4",
"Find y(5) for y(x) = max{x, 30-x}",
"Find the value of t(h) where h=-15 if t(h)=sqrt(81-h^2)+h/h-5. The value can be complex.",
"Evaluate f(-7) if f(m) = m*(m-6)*(m-12)+(m-7)^2+(m+2)^2",
"Let c(g) = 23*g + 31, and let d(f) = f + 41, find c(d(12))",
"For x(g) = g/g-9+11/g, determine the value of x(6)",
"compute the average value of f(theta) over 0 to 2pi for f(theta) = sin^2 theta + cos^2 2theta",
"If t(s) = s/(s-6), find the value of t(t(t(2)))"
\end{minted}